\documentclass[journal]{IEEEtran}

\usepackage{graphicx} 
\usepackage[colorlinks,linkcolor=blue,citecolor=blue]{hyperref}
\usepackage{enumitem}
\usepackage{amsmath}
\usepackage{amssymb}
\usepackage{multirow}
\usepackage{colortbl}
\usepackage[normalem]{ulem}
\usepackage[switch]{lineno}

\ifCLASSINFOpdf
\else
\fi

\begin{document}

\title{WaveInst: A Frequency-Domain Enhanced Network for Fine-Grained Thin Tree Trunk Extraction in  Forest Scenes}

\author{Chenyang~Fan, Xujie~Zhu, Taige~Luo, Zhulin~Chen, and Sheng~Xu

\thanks{This work was supported in part by the Biological Breeding-National Science and Technology Major Project (NO.2023ZD0405605), and in part by the Practice Innovation Training Program Projects for Jiangsu College Students under Grant 202510298056Z.\textit{(Corresponding author: Sheng Xu.)}}
\thanks{Chenyang Fan, Xujie Zhu, and Sheng Xu are with College of Information Science and Technology \& Artificial Intelligence, Nanjing Forestry University, Nanjing 210037, China (e-mail: fancy@njfu.edu.cn; xujiezhu@njfu.edu.cn; xusheng@njfu.edu.cn).}
\thanks{Taige Luo is with College of Information Science and Technology \& Artificial Intelligence, Nanjing Forestry University, Nanjing 210037, China, and also with Department of Geography, College of Natural Resources and Environment, Virginia Tech, Blacksburg 24061, USA (e-mail: taige@vt.edu).}
\thanks{Zhulin Chen is with the Institute of Forest Resource Information Techniques, Chinese Academy of Forestry, Beijing 100091, China, and also with State Forestry and Grassland Administration, Key Laboratory of Forest Management and Growth Modelling, Beijing 100091, China (e-mail: chenzl@ifrit.ac.cn).}}

\markboth{Journal of \LaTeX\ Class Files,~Vol.~14, No.~8, August~2015}%
{Shell \MakeLowercase{\textit{et al.}}: Bare Demo of IEEEtran.cls for IEEE Journals}

\maketitle

\begin{abstract}
Analyzing tree morphology, particularly trunk and branch extraction, is valuable for genetic breeding and forestry management. Existing image-based deep learning methods tend to misidentify overlapping trunks as a single trunk when structural discontinuities occur due to front-back overlap, while low contrast between trunk textures and the background further complicates segmentation. Moreover, limited juvenile tree data, coupled with substantial variations in trunk diameter across growth stages, restricts model performance in extracting thin trunks and branches. Based on this, we propose an instance segmentation network leveraging frequency-domain features, WaveInst. Its core is a Frequency-domain Feature Compensation branch, consisting of Discrete Wavelet Transform block and High-Frequency Enhancement block. The former performs high- and low-frequency decomposition and aggregates high-frequency responses along multiple directions, while the latter further refines the high-frequency features through multi-path processing. An Adaptive Gated Fusion Module is then applied to effectively integrate spatial-domain convolutional features with frequency-domain representations, allowing the decoder to utilize embedded frequency-domain features to enhance its fine-grained detail representation. We conduct experiments on public datasets including SynthTree43k, CaneTree100, and UrbanStreet, as well as PoplarDataset, which contains both mature and juvenile poplar trees. On the public datasets, WaveInst demonstrates strong performance and stable robustness across diverse scenarios. On PoplarDataset, it achieves a mean average precision of 53.1 for mature and 29.1 for juvenile, outperforming existing state-of-the-art methods on juvenile by 6.6 points, demonstrating its effectiveness in extracting thin tree trunk.
\end{abstract}


\IEEEpeerreviewmaketitle

\section{Introduction}

\IEEEPARstart{I}{n} modern forestry, scientific breeding stands as a key pillar for enhancing forestry productivity. By utilizing advanced phenotyping technologies for precise measurement and analysis, it offers robust scientific underpinning for variety improvement and breeding decision-making \cite{pieruschka2019plant, bian2022closing, zhang2023crop}. Among various phenotypic traits, the structural characteristics of individual trees are particularly crucial, as they not only reflect growth and health status but also provide a foundational basis for data-driven forest management \cite{vuidot2011influence}. It is through such in-depth exploration of individual tree structure that we can unlock new potential for more effective and efficient forest management, thereby elevating the entire realm of forestry production to new heights.

Current research on tree structure analysis is based on point clouds and unmanned aerial vehicle (UAV) imagery. Point clouds provide high-precision and high-dimensional geometric information, which is essential for the 3-D reconstruction and comprehensive analysis of tree structures \cite{wang2023branching, gao2025extraction, yun2025framework}. By using point clouds, detailed structural data of trees can be captured with high accuracy, offering a solid foundation for the development of tree growth models and phenological monitoring. However, the high cost of light detection and ranging (LiDAR) equipment restricts its widespread application in large-scale forest surveys and growth monitoring \cite{kuang2024comprehensive}. Additionally, the large volume of data collected is frequently contaminated by noise, complicating subsequent processing and increasing the overall computational load. The substantial demand for computing resources further hinders the efficiency of real-time data processing and application \cite{wang2025breaking}.

In contrast, UAV-based technologies offer a cost-effective and flexible alternative, enabling rapid acquisition of high-resolution imagery across large forest areas using multispectral or visible light sensors. Moreover, UAV imagery can be integrated into temporal monitoring frameworks to track the dynamics of tree growth over time, providing continuous data for long-term phenotypic studies \cite{d2020high, firoze2023tree}. However, in natural forest environments, the quality and consistency of UAV imagery can be affected by various external factors. These include canopy occlusion, variable illumination caused by weather or time of day, and heterogeneous backgrounds due to dense understory vegetation or terrain variation. Such variability poses challenges for maintaining stable image quality and limits the generalizability of downstream phenotypic analysis.

\begin{figure}[h]
	\centering
	\includegraphics[width=\linewidth]{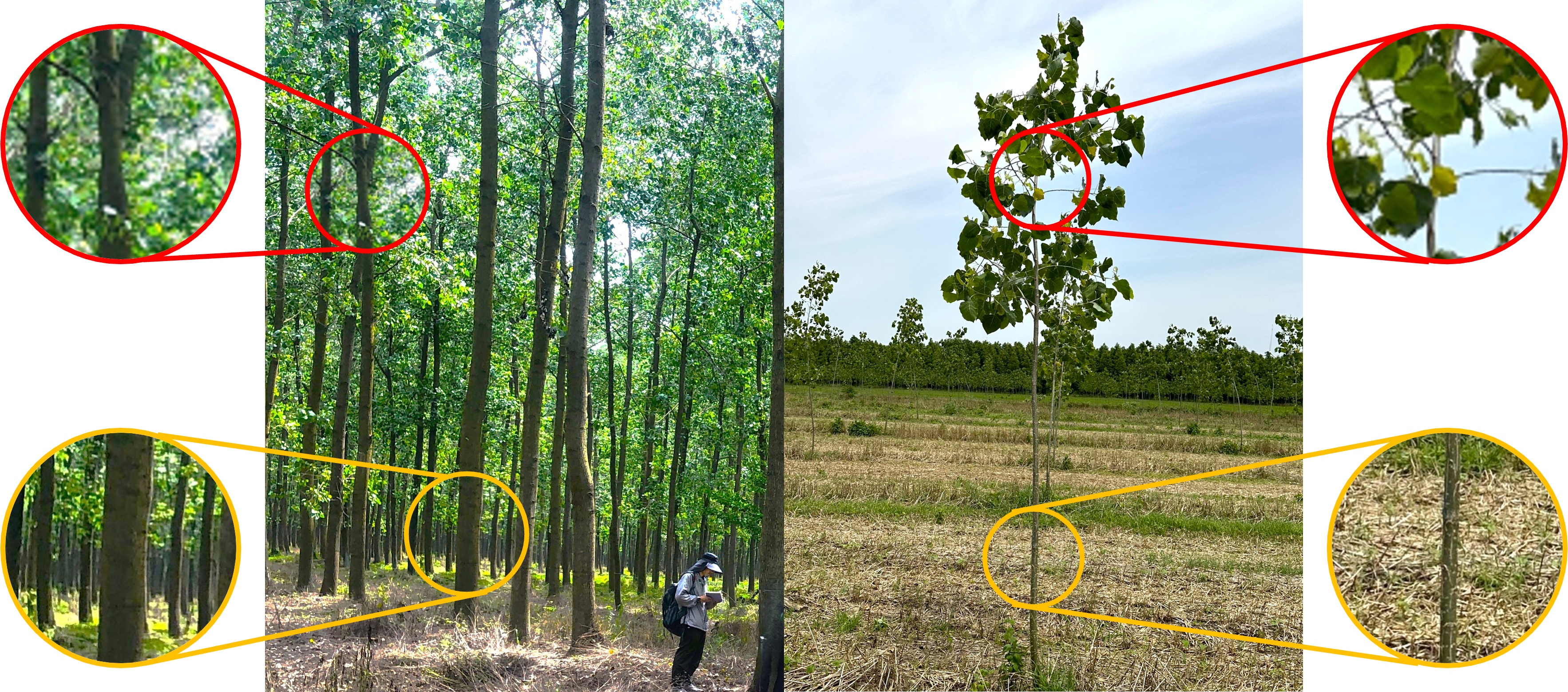}
	\caption{Challenges of tree trunk structure extraction. Structural discontinuities caused by occlusions (red), along with trunk thickness variation and low contrast between tree textures and background (yellow).}
	\label{fig:1}
\end{figure}

Existing methods for tree trunk extraction based on UAV imagery primarily include object detection and image segmentation \cite{alaya2021cnn, jodas2021deep, li2023tree}. Object detection methods can locate the trunk's boundary, but the generated bounding boxes often contain background information, which is difficult for structural phenotypic measurement. Image segmentation allows for pixel-level prediction, but it is vulnerable to interference in complex natural environments. As illustrated in Fig. \ref{fig:1}, occlusions, trunk thickness variation, and the weak contrast between juvenile tree textures and their backgrounds often lead to structural discontinuities or inaccurate segmentation. These challenges are especially pronounced for younger trees, contributing to a research gap compared to mature tree analysis \cite{fraser2018issues, hui2022multi}.

To address these challenges, we propose an instance segmentation approach that integrates wavelet transform with deep neural networks. Specifically, we introduce a novel Discrete Wavelet Transform (DWT) Block and a High-Frequency Enhance (HFE) Block to perform multi-scale edge enhancement, effectively preserving and reinforcing spatial detail information. This design mitigates the issues of edge blurring and detail loss commonly seen in traditional segmentation methods, enabling accurate extraction of individual tree structures from UAV imagery. The resulting outputs provide more precise and reliable data for downstream phenotypic analysis at the individual tree level. 

The main contributions of this article are listed as follows.

\begin{enumerate}[label=\arabic*)]
    \item We propose a novel refine segmentation framework for the tree structure extraction, with a wavelet-based high-frequency feature compensation branch to achieve multi-scale feature maps for overcoming branch blurring and detail loss in forest vegetation imagery.
    \item We design a High-Frequency Enhance block to enhance the discriminative information in the high-frequency sub-bands for selectively amplifying critical edge and texture features in the small branches of the seedlings.
    \item We introduce an Adaptive Gated Fusion Module is to integrate spatial-domain convolutional features with frequency-domain representations for improving the continuity of trunks in the area of leaf occlusion.
\end{enumerate}

The remainder of this article is organized as follows. Section \ref{sec:2} provides an overview of the related works on individual tree structure extraction. In Section \ref{sec:3}, the architecture of the proposed WaveInst is introduced. Section \ref{sec:4} presents a comprehensive description of the extensive experiments conducted. Section \ref{sec:5} discusses runtime analysis and applications for estimating DBH, tree height, and recognizing tree species. Finally, Section \ref{sec:6} concludes the article.

\section{Related works}
\label{sec:2}

The study of individual tree structures typically focuses on components such as the canopy, trunk, and branches, aiming to accurately extract morphological information to support applications in forestry resource assessment, ecosystem monitoring, and growth modeling. Existing methods can be broadly categorized into traditional image processing techniques based on handcrafted features, and deep learning approaches that learn features automatically from data.

\subsection{Traditional Image Processing Approaches}

Image processing techniques have historically played a central role in tree structure analysis, relying on manually crafted features and heuristic rules. These methods typically involve steps such as image preprocessing, feature extraction, segmentation, and classification. Erikson and Olofsson \cite{erikson2005comparison} systematically summarized commonly used individual tree crown detection methods, such template matching, segmentation supported by fuzzy rules and segmentation by Brownian motion, showing that sm all tree crowns are difficult to detect due to confusion with the background. Subsequently, Huertas et al. \cite{huertas2005stereo} proposed a stereo vision–based forest traversability analysis method, which constructs spatial models of tree trunks using depth maps and evaluates traversability. This method demonstrated high robustness in real forest environments, though its accuracy decreased under trunk occlusion or complex background conditions. Lu and Rasmussen \cite{lu2011tree} designed a multi-scale bar-shaped filter based on contrast differences, enabling tree trunk localization through vertical edge detection, but its adaptability to complex backgrounds requited to be improved. Overall, such traditional approaches rely on hand-crafted features or thresholds and scene-specific prior knowledge, resulting in limited generalization ability in complex environments \cite{anubha2019study}.

\subsection{Deep Learning Approaches}

In recent years, deep learning has demonstrated superior robustness over traditional methods in individual tree structure analysis by automatically learning features from large-scale datasets \cite{yun2024status}. In individual tree crown structure extraction, Luo et al. \cite{luo2024vrsnet} proposed a Gaussian isocontour-based density map generation algorithm, which utilizes embedded semantic information to achieve precise crown segmentation and counting, demonstrating the potential of density modeling in forestry target detection. In addition, Wen et al. \cite{wen2024comparing} compared the performance of Mask R-CNN and a traditional watershed-based method for individual tree crown segmentation, showing that the deep learning approach outperforms in both accuracy and efficiency. Their results showed that deep learning improved accuracy and adaptability in complex forest canopies.

\begin{figure*}[h]
	\centering
	\includegraphics[width=\textwidth]{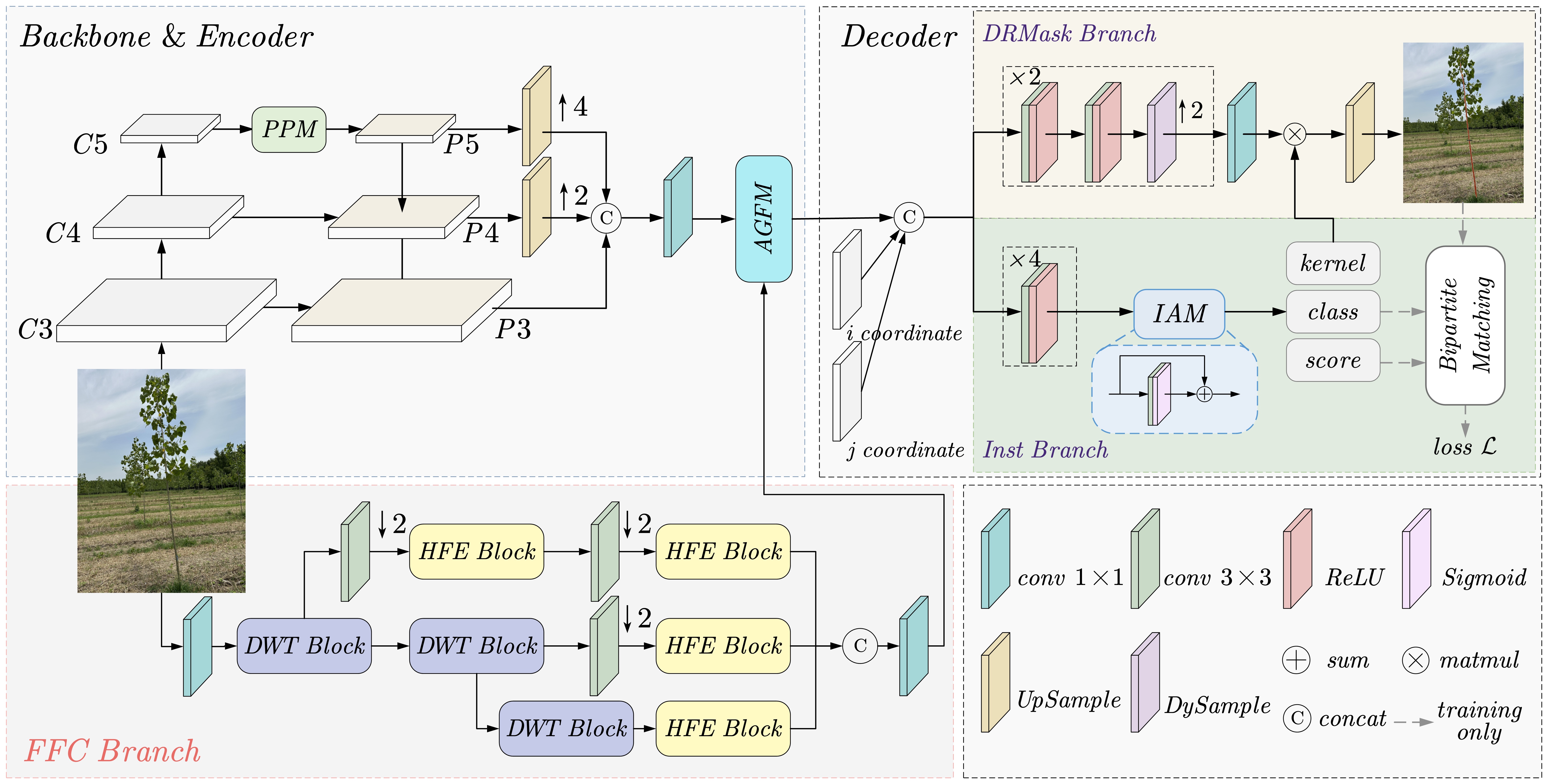}
	\caption{Overall architecture of the proposed WaveInst.}
	\label{fig:2}
\end{figure*}

For tree trunk structure extraction under the canopy, Silva et al. \cite{silva2022line} proposed a line-based deep learning method that combines High-Resolution Network (HRNet) with a line structure regression branch, enabling accurate detection of tree branches. By explicitly modeling the linear structure of branches, this method significantly improves detection performance in complex backgrounds, particularly in localized areas where branch shapes are clear but easily occluded. However, its applicability is primarily limited to local structure extraction, which makes it less suitable for modeling the global structure of entire trees. To address this, Mo et al. \cite{mo2025real} proposed the RT-Trunk network to meet the real-time demands of trunk extraction in understory environments. Based on an instance segmentation framework, the method enables efficient extraction of individual tree trunks. However, their research mainly focused on mature trees, with limited adaptability to saplings, primarily due to the lack of sapling datasets and the labor-intensive nature of dataset annotation. To improve the generalization ability of models to unseen tree species, large-scale pre-trained models and zero-shot learning mechanisms have been introduced. Yu et al. \cite{yu2025efficient} proposed the SGS-3DRecon framework, which integrates Segformer, GAN inverse mapping, and Sobel operator for image-based tree trunk volume estimation. The core method involves using Segformer for semantic segmentation to obtain the tree trunk mask and employing GAN to restore the mask of occluded areas. However, this process often generates jagged artifacts due to upsampling and downsampling operations. To address this issue, the authors introduced the Sobel operator for boundary smoothing, optimizing the boundary regions and effectively reducing the artifacts. Osco et al. \cite{osco2023segment} incorporated Meta AI’s Segment Anything Model (SAM) into remote sensing-based tree instance segmentation tasks to evaluate its performance in identifying canopy and trunk structures in forest scenes. Although SAM was not specifically trained on forestry data, it demonstrated a certain degree of structural recognition capability when guided with appropriate prompts, offering a novel approach for few-shot and even zero-shot tree structural detection.

Although existing methods have achieved promising results in tree structure extraction, several challenges remain. Repeated downsampling in deep networks often suppresses high-frequency signals, leading to the loss of ultra-fine structures such as sapling branches. Moreover, background suppression mainly relies on semantic features in the image domain, which become less discriminative when the contrast between young trees and surrounding vegetation is weak. In addition, most models lack adaptive mechanisms to handle large variations in trunk thickness across growth stages. To address these issues, this study introduces a frequency-enhanced framework in which a DWT Block in the FFC Branch captures high-frequency structural cues, an HFE Block strengthens edge responses, and an AGFM integrates spatial and frequency-domain features for more robust structure extraction.

\section{PROPOSED METHOD}
\label{sec:3}

\begin{figure*}[h]
	\centering
	\includegraphics[width=\textwidth]{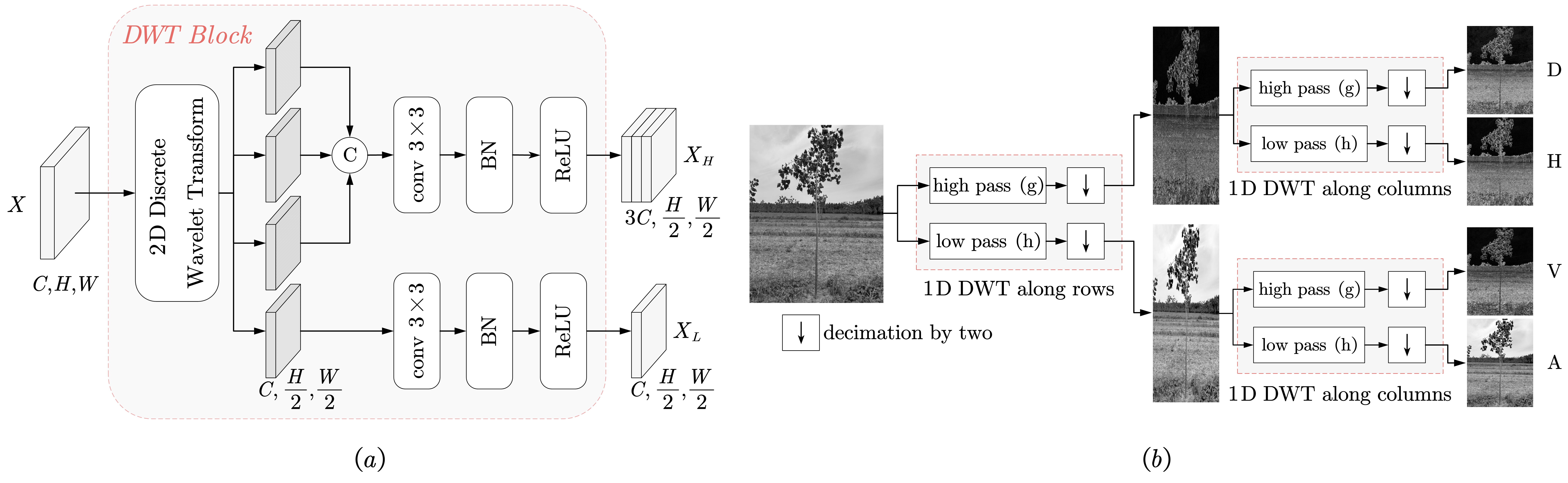}
	\caption{(a) Structure of DWT Block; (b) Principles and results of the Discrete Wavelet Transform.}
	\label{fig:3}
\end{figure*}

The proposed network, WaveInst, is designed for fine-grid extraction of tree trunks from understory images, enabling precise and detailed structural segmentation. As shown in Fig. \ref{fig:2}, it consists of four main components: the backbone, the encoder, the decoder, and a Frequency-domain Feature Compensation (FFC) branch. The FFC branch is composed of a Discrete Wavelet Transformation (DWT) Block and a High-Frequency Enhancement (HFE) Block. The encoder aggregates multi-scale features extracted by the backbone through a Feature Pyramid Network (FPN) and utilizes the proposed Adaptive Gated Fusion Module (AGFM) to adaptively fuse semantic and frequency-domain information. The decoder employs CoordConv and a decoupled dual-branch design: the DRMask branch reconstructs fine spatial details and generates segmentation masks, while the Inst branch discriminates instances and refines instance-level predictions. Finally, a bipartite matching mechanism ensures precise supervision for accurate instance segmentation.

\subsection{Frequency-domain Feature Compensation (FFC) Branch}

Current convolutional neural networks typically reduce spatial resolution and increase channel dimensions during feature extraction to lower computational overhead. However, aggressive downsampling often violates the Nyquist–Shannon sampling theorem, resulting in spectral aliasing \cite{vasconcelos2020effective}. When the sampling rate falls below twice the highest frequency component, high-frequency information is folded into lower bands, causing edge artifacts and texture distortions. These distortions make it difficult for the network to recover fine structural details in subsequent layers, ultimately limiting segmentation accuracy.

To address this issue, we propose a frequency-domain feature compensation (FFC) branch based on discrete wavelet decomposition. Operating in parallel with the backbone network, this branch leverages multiresolution analysis in the wavelet domain to decouple frequency bands while preserving the efficiency of spatial downsampling. Specifically, it employs an orthogonal filter bank to extract high-frequency details in a directional manner and embed high-frequency priors into the network. The DWT Block performs multiscale decomposition, effectively separating low- and high-frequency components, while the HFE Block selectively enhances high-frequency subbands to enrich edge and texture information. This design compensates for the loss of fine details during downsampling, providing richer structural cues and improving segmentation precision.

\subsubsection{Discrete Wavelet Transformation (DWT) Block}

This block is designed to perform frequency-domain decomposition of the input feature maps and aggregate directional high-frequency information while maintaining the efficiency of spatial downsampling. Specifically, a 2D discrete wavelet transform is applied along both the row and column directions. We adopt the Haar (db1) wavelet in our model for its efficiency and effectiveness in capturing fine high-frequency structures, with low-order wavelets providing sharp edge responses useful for thin tree trunk segmentation. The specific effects of wavelet basis selection and wavelet order are discussed in detail in Section \ref{sec:ablation}. The decomposition uses a chosen wavelet basis $\psi$ and scaling function $\phi$ to construct high-pass and low-pass filters. For an input signal $x$, the row-wise filtering can be expressed as:
\begin{align}
    L_r[i,j] = \sum_{k} x[i,k] \, \phi[k - 2j], \\
	H_r[i,j] = \sum_{k} x[i,k] \, \psi[k - 2j]
\end{align}

where $L_r$ and $H_r$ are the row-wise low-frequency and high-frequency components, respectively. Similarly, column-wise filtering is applied to $L_r$ and $H_r$ to obtain the four subbands:
\begin{align}
	D &= H_c(H_r), & H &= H_c(L_r), \nonumber \\
	V &= L_c(H_r), & A &= L_c(L_r)
\end{align}

Here, $D$, $H$, and $V$ represent high-frequency components in diagonal, horizontal, and vertical directions, respectively, while $A$ is the low-frequency approximation containing rich global information. To better embed high-frequency priors and guide the network to focus on the boundaries between tree trunks and background, the high-frequency subbands $(D, H, V)$ are concatenated and passed through a $3\times3$ convolution with ReLU activation to aggregate directional details, yielding the joint high-frequency feature:
\begin{align}
	X_H = \mathrm{ReLU}\Big(\mathrm{BN}\Big( \mathrm{conv}_{3\times3}\big(\mathrm{concat}(D,H,V)\big) \Big)\Big)
\end{align}

For the low-frequency subband $A$, feature extraction is performed followed by a nonlinear activation to produce the low-frequency feature:
\begin{align}
	X_L = \mathrm{ReLU}\Big( \mathrm{BN}\Big( \mathrm{conv}_{3\times3}(A)\Big) \Big)
\end{align}

This low-frequency feature $X_L$ preserves global structural information and smooth contextual cues, and is used for further frequency-domain decomposition to enable multi-scale representation.

\subsubsection{High-Frequency Enhancement (HFE) Block}

\begin{figure}[h]
	\centering
	\includegraphics[width=\linewidth]{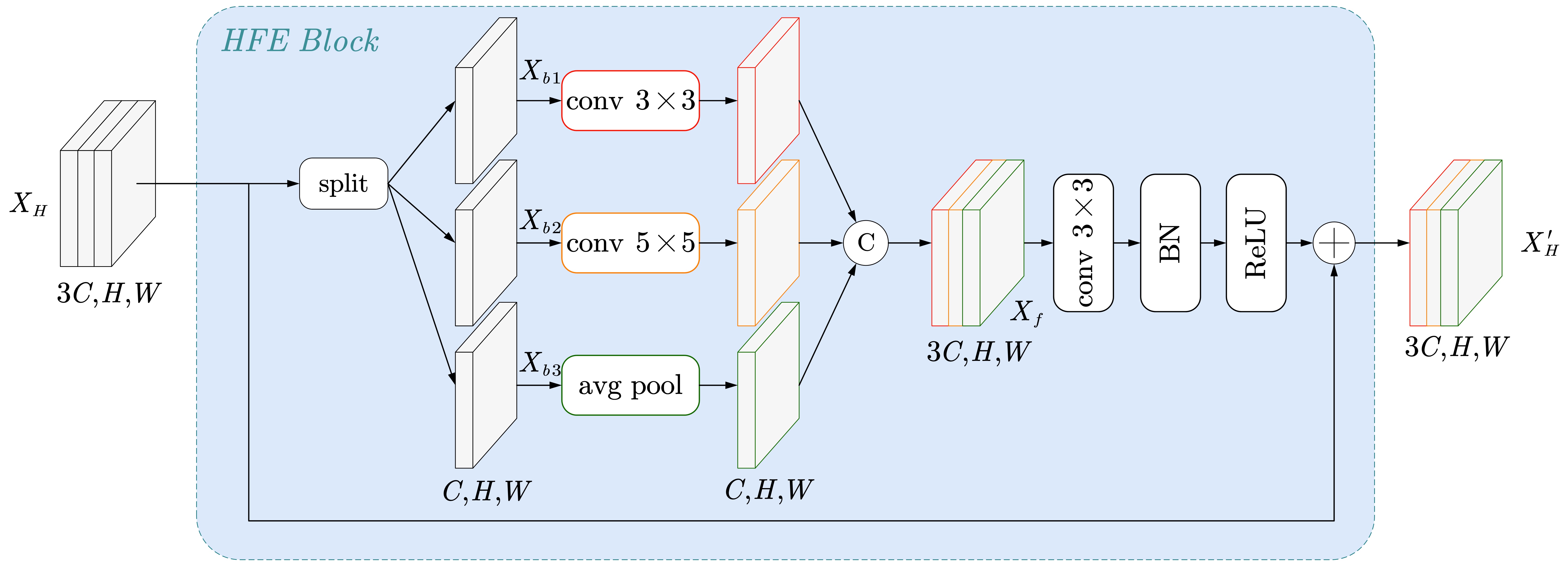}
	\caption{Structure of HFE Block.}
	\label{fig:4}
\end{figure}

The HFE Block further refines the high-frequency feature $X_H$ generated by the DWT Block. As illustrated in Fig. \ref{fig:4}, it employs parallel feature extraction and residual connections to capture fine-grained details.  

Considering that in UAV-captured understory images, trees predominantly grow in a vertical orientation, the boundaries between the tree trunks and the background typically lie in the horizontal and diagonal directions. Given that wavelet-decomposed sub-bands contain different frequency and directional information, we applied different processing strategies to these sub-bands based on the specific information they contain, which helps capture and express the boundaries more effectively.

Specifically, the high-frequency feature $X_H$ is first split along the channel dimension into three blocks corresponding to the diagonal, horizontal, and vertical directions. A 3×3 convolution kernel is used for the diagonal direction to capture fine, localized feature variations, while a 5×5 convolution kernel is applied to the horizontal direction to capture broader and more stable tree trunk boundaries. In contrast, the vertical direction, where the trunk structure is more stable, benefits from average pooling, which helps smooth texture variations and reduce noise, allowing the model to focus on the overall structure of the tree trunk. Following this, the features extracted from different directions are concatenated along the channel dimension to obtain the enhanced high-frequency response feature $X_f$. This multi-path processing strategy, compared with a single unified convolution, generates richer frequency-domain feature maps and effectively aggregates complementary information from receptive fields of different sizes. The expression is as follows:
 
\begin{align}
	X_f = \mathrm{concat}\Big(
	&\mathrm{conv}_{3\times3}(X_{b1}), \ 
	\mathrm{conv}_{5\times5}(X_{b2}), \nonumber \\
	&\mathrm{avgpool}_{3\times3}(X_{b3})
	\Big)
\end{align}

Here, the concatenated feature $X_f$ is further fused using a $3\times3$ convolution layer, followed by a ReLU activation to introduce non-linearity. A residual connection is added to enhance expressiveness and mitigate gradient vanishing, producing the refined high-frequency feature $X_H'$:  
\begin{align}
	X_H' = \mathrm{ReLU}\Big(\mathrm{BN}\Big(\mathrm{conv}_{3\times3}(X_f)\Big)\Big) + X_H
\end{align}

This design allows the network to effectively aggregate directional high-frequency details while preserving the original high-frequency information, improving both the precision of boundary representation and the efficiency of training.

\subsection{Adaptive Gated Fusion Module (AGFM)}

We design a lightweight Adaptive Gated Fusion Module (AGFM) to adaptively fuse the multi-scale semantic information from the FPN with the spectral features from the FFC Branch. Specifically, in the original encoder, the multi-scale feature maps $P3 \sim P5$ produced by the FPN are upsampled to the size of $P3$ using bilinear interpolation, concatenated along the channel dimension, and then compressed along the channel dimension to form the joint multi-scale semantic feature map, denoted as $F_{fpn}$. Correspondingly, the high-frequency feature map from the FFC Branch is denoted as $F_{dwt}$, which retains more edge and texture details.

\begin{figure}[h]
	\centering
	\includegraphics[width=\linewidth]{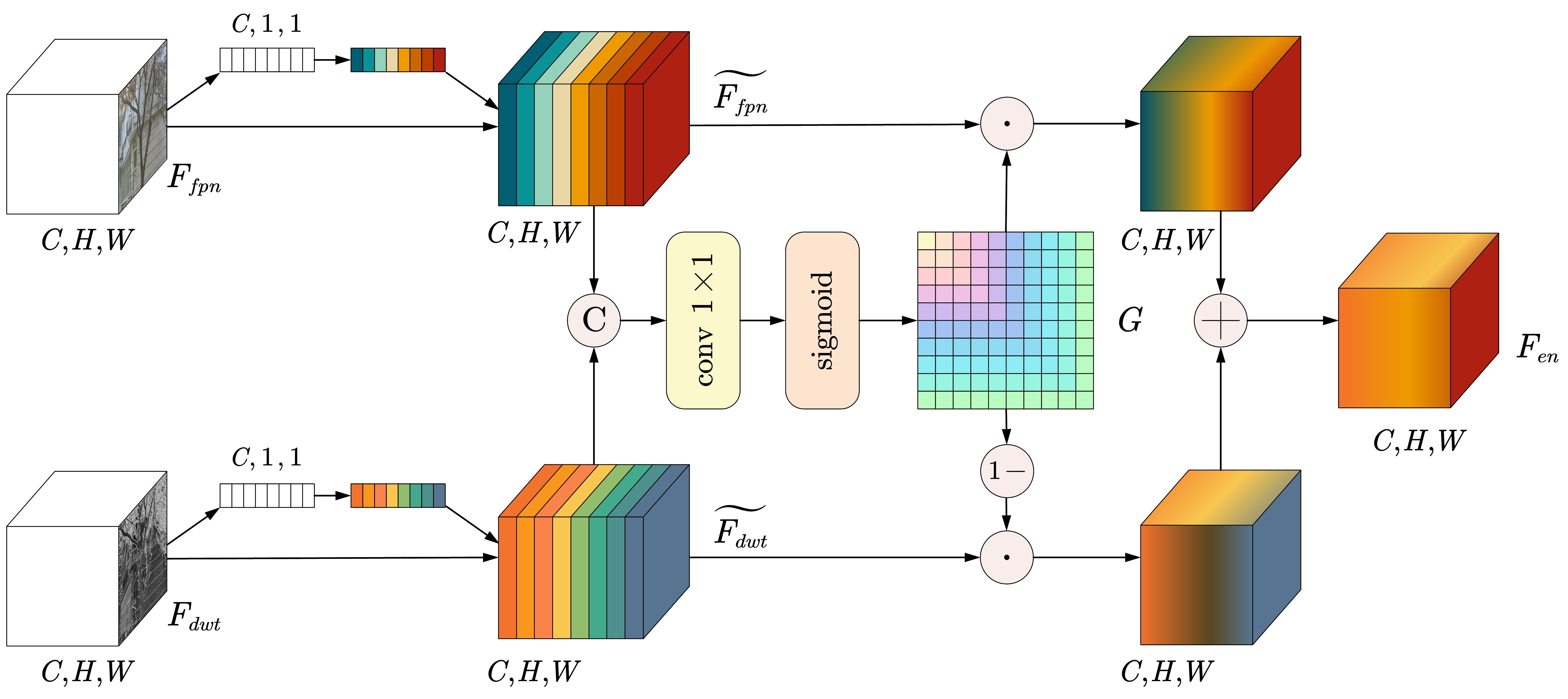}
	\caption{Structure of Adaptive Gated Fusion Module (AGFM).}
	\label{fig:5}
\end{figure}

As illustrated in Fig. \ref{fig:5}, the computation process of the AGFM is as follows. To enhance the channel-wise importance within each branch, we apply a Squeeze-and-Excitation (SE) channel attention mechanism to both $F_{fpn}$ and $F_{dwt}$, which adaptively emphasizes key channel responses and suppresses redundant information \cite{hu2018squeeze}. 
The resulting enhanced feature maps are denoted as $\widetilde{F_{fpn}}$ and $\widetilde{F_{dwt}}$.

Subsequently, the two attention-enhanced feature maps are concatenated along the channel dimension 
to form a joint feature tensor of size $(2C,H,W)$. This tensor is then passed through a $1\times1$ gated convolution layer, followed by a sigmoid activation function, to generate a spatial 
gating map $G \in \mathbb{R}^{1 \times H \times W}$. The computation of $G$ can be formally expressed as:
\begin{align}
	G = \sigma \Big( \text{conv}_{1\times1} \big( \text{concat}(\widetilde{F_{fpn}}, \widetilde{F_{dwt}}) \big) \Big),
\end{align}

where $\sigma$ denotes the sigmoid function, which maps the fused feature values to the $[0,1]$ range, 
producing a weight map that adaptively modulates the contribution of features from each branch at every spatial location.

During the fusion process, $\widetilde{F_{fpn}}$ is element-wise multiplied by $G$, while $\widetilde{F_{dwt}}$ is multiplied by $1-G$, generating two weighted feature maps. The final fused feature map $F_{en}$ is then obtained through element-wise addition:
\begin{align}
	F_{en}=G\odot \widetilde{F_{fpn}}+\left( 1-G \right) \odot \widetilde{F_{dwt}}
\end{align}

This design allows the AGFM to adaptively learn the relative importance of features from the FPN and FFC branch, enhancing discriminative information while preserving fine-grained edge and texture details.

\subsection{DySample Module}

In the field of instance segmentation, existing models generally use feature maps with a resolution of 1/4 or 1/8 of the original image size to construct mask segmentation heads, which are then restored to the original image size through bilinear interpolation. While this approach balances computational efficiency and segmentation performance, it has significant drawbacks when applied to fine-grained plant phenotype analysis. The use of low-resolution feature maps results in the loss of high-frequency details, leading to suboptimal segmentation of branch edges, particularly in the analysis of juvenile trees. Furthermore, traditional interpolation methods are limited by the absence of local prior knowledge and fixed kernel weights, making it challenging to accurately reconstruct complex textures and fine structures, ultimately impacting the precision of the segmentation results \cite{zhao2017iterative}.

To improve segmentation accuracy and preserve critical details, we introduce a lightweight dynamic sampling module, DySample, into the DRMask (Dynamic Reconstruction Mask) branch. The feature map is restored to half the original image size before mask generation. By progressively reconstructing high-resolution feature maps, the segmentation results are refined over multiple stages, effectively avoiding the loss of details and computational overhead that can occur in single-stage reconstruction, while better capturing local features within the image \cite{lai2017deep, wang2018fully}. Unlike traditional interpolation methods, DySample leverages learnable dynamic offsets to guide the network in autonomously identifying optimal sampling points within a continuous feature space, thus enabling semantic-aware feature reconstruction \cite{liu2023learning}.

\begin{figure}[h]
	\centering
	\includegraphics[width=\linewidth]{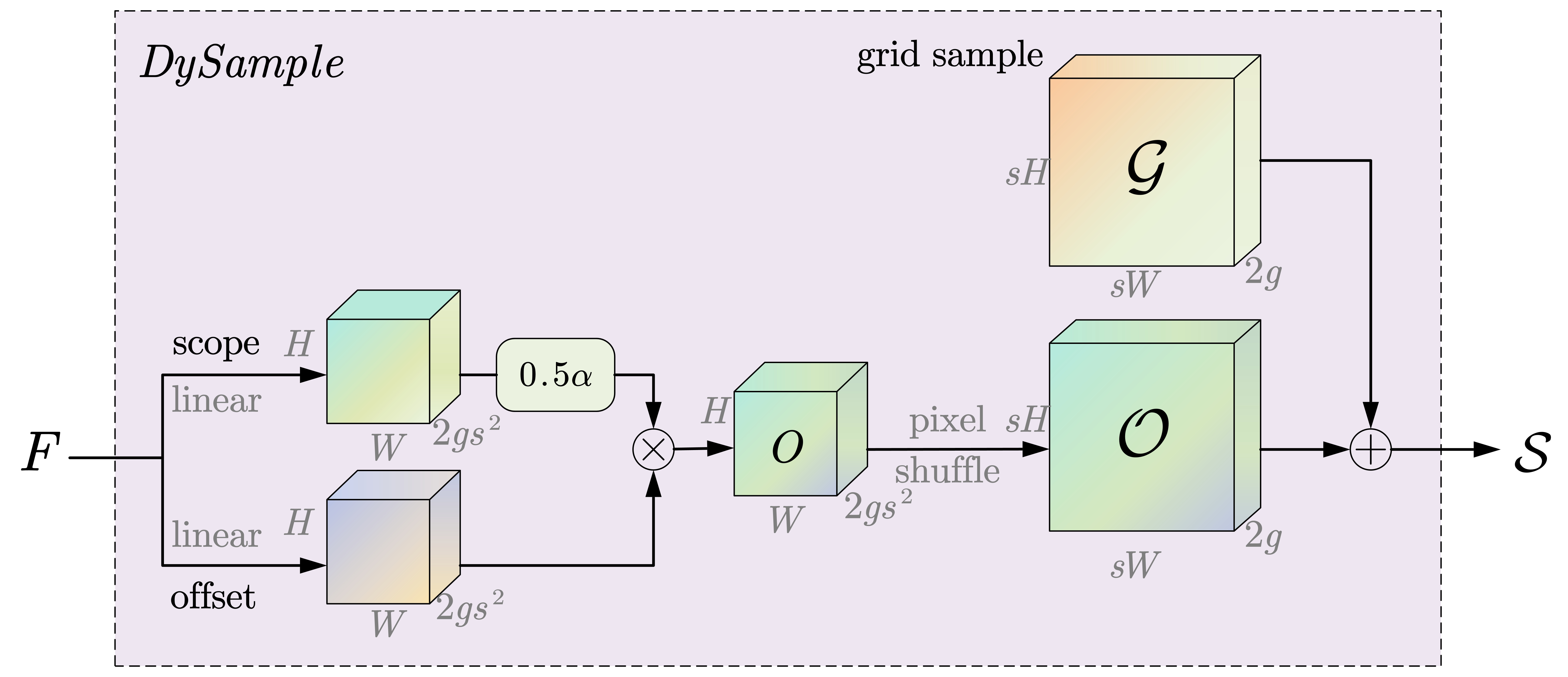}
	\caption{Structure of DySample Module.}
	\label{fig:6}
\end{figure}

Fig. \ref{fig:6} illustrates the dynamic feature reconstruction process of the DySample module, which primarily achieves semantically adaptive feature reconstruction through grouped linear projection and dynamic offset constraints. Specifically, given an input feature map $F\in \mathbb{R} ^{C\times H\times W}$, two parallel $1\times1$ linear projections are first applied to generate feature matrices of shape $(2gs^2, H, W)$, where $g$ represents the number of groups and $s$ denotes the upsampling factor, ensuring that the sampling resolution aligns with the final reconstruction requirements. Among these two projection branches, the lower branch is responsible for generating the dynamic offset field, while the upper branch modulates the offset magnitude. The output of the upper branch undergoes sigmoid normalization and is scaled by a factor of $0.5$ before being combined with the offset field from the lower branch using the Hadamard product. This operation dynamically constrains the offset range, ensuring smoother and more stable sampling position adjustments. The process can be mathematically formulated as:
\begin{align}
	O = \text{offset}(F) \circ \sigma(\text{scope}(F)) \times 0.5
\end{align}
where $\sigma(\cdot)$ and $\circ$ represent the sigmoid function and the Hadamard product, respectively.

\begin{figure*}[h]
	\centering
	\includegraphics[width=\textwidth]{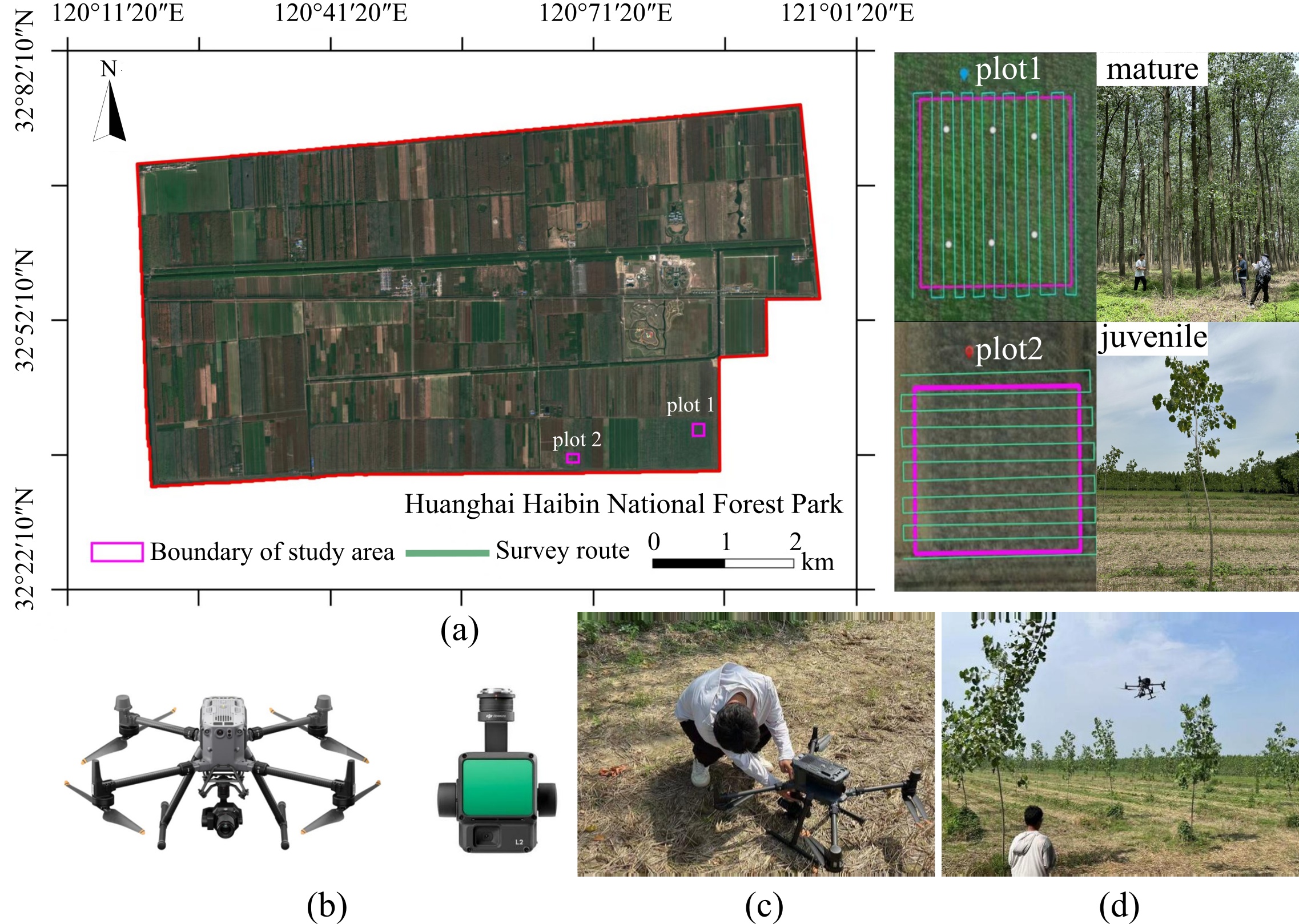}
	\caption{Study areas and data acquisition. (a) The site of the poplar plantation within Huanghai Haibin National Forest Park, including the scale of the study area. (b) UAV imagery collection equipment: M350RTK. (c-d) The process of imagery collection.}
	\label{fig:7}
\end{figure*}

Subsequently, the offset field $O$ is refined through Pixel Shuffle, which reorganizes its spatial structure to align with the upsampling ratio, yielding the final offset field:
\begin{align}
	\mathcal{O} =\mathrm{PixelShuffle}(O)
\end{align}
This operation maps the learned offsets to the resolution feature space corresponding to the upsampling factor $s$. The final adaptive sampling set $\mathcal{S}$ is obtained by adding the offset field $\mathcal{O}$ to the initial sampling grid $\mathcal{G}$, i.e.:
\begin{align}
	\mathcal{S} = \mathcal{G} + \mathcal{O}
\end{align}
where $\mathcal{G}$ is the initial sampling grid. In simple terms, this means that the final sampling points are determined by adding offsets to the original grid, allowing for more precise capturing of the features or regions that need attention.

\subsection{Bipartite Matching and Loss}

In our work, the instance segmentation task is redefined through task decoupling into two synergistically optimized sub-tasks: Mask Geometry Modeling and Mask-Semantic Correspondence \cite{cheng2021per, xie2021segformer}. This concept is fully embodied in our proposed WaveInst, where the decoder adopts a dual-branch architecture: the DRMask branch is responsible for pixel-wise mask prediction, while the Inst branch generates instance-level semantic information based on the instance activation maps (IAM) and establishes the correspondence between masks and semantic labels through bipartite matching.

We project the instance activation maps $A \in \mathbb{R}^{N \times \frac{H}{8} \times \frac{W}{8}}$ into the semantic space through a learnable class linear layer $W \in \mathbb{R}^{N \times C}$ (where $N = 100$ denotes the number of predicted instances and $C$ represents the number of categories), thereby obtaining the category distribution probabilities for each predicted instance. Then, the matching cost for the $i$-th prediction and the $k$-th ground-truth object is defined as:
\begin{align}&&
	Cost(i,k) = p_{i,c_{k}}^{1-\alpha} \cdot \mathrm{DICE}(m_{i},g_{k})^{\alpha}
	&\end{align}
where $\alpha$ is a hyperparameter used to balance the impact of classification and segmentation, and it is empirically set to 0.8. $c_{k}$ is the category label for the $k$-th ground-truth object. $p_{i,c_{k}}$ represents the probability of the $i$-th prediction belonging to class $c_{k}$, which is obtained from the class linear layer. $m_{i}$ and $g_{k}$ are the masks for the $i$-th prediction and the $k$-th ground-truth, respectively. The dice score between the prediction and the ground-truth is defined as:
\begin{align}
	\mathrm{DICE}(m,g) = \frac{2\sum\nolimits_{x,y} m_{xy} \cdot g_{xy}}{\sum\nolimits_{x,y} m_{xy}^2 + \sum\nolimits_{x,y} g_{xy}^2}
\end{align}
where $m_{xy}$ and $g_{xy}$ represent the pixel values at position $(x, y)$ in the predicted mask $m$ and the ground-truth mask $g$, respectively. We formulate the global optimal bipartite matching between $N$ predicted instances and $K$ ground-truth objects based on a dynamically constructed matching score matrix $Cost$, using the Hungarian Algorithm \cite{stewart2016end}. This algorithm establishes one-to-one correspondences by maximizing the bidirectional matching scores, which can be mathematically formalized as:
\begin{align}
	\pi^{\ast} = \underset{\pi \in \Pi(N, K)}{\mathrm{arg}\max} \sum\nolimits_{N,K} Cost(i,k) \cdot \mathbb{I}_{\left\{ \pi(i) = k \right\}}
\end{align}
where $\Pi(N, K)$ represents the set of all bijective mappings between the $N$ predictions and the $K$ ground-truths, and $\mathbb{I}(\cdot)$ is the indicator function.

To optimize the learning process of the instance segmentation task, we design a comprehensive loss function, as shown in Equation \ref{eqs:loss}, which integrates four components: classification loss, objectness loss, dice loss, and pixel-wise mask loss.

\begin{align}
	\label{eqs:loss}
	\mathcal{L} = \lambda_{cls} \cdot \mathcal{L}_{cls} + \lambda_{obj} \cdot \mathcal{L}_{obj} + \lambda_{dice} \cdot \mathcal{L}_{dice} + \lambda_{pix} \cdot \mathcal{L}_{pix}
\end{align}
where, $\mathcal{L}_{cls}$ adopts focal loss (with $\alpha=0.25$, $\gamma=2.0$) to address class imbalance by dynamically re-weighting uncertain samples, thereby improving model robustness \cite{lin2017focal}. The objectness loss $\mathcal{L}_{obj}$ is computed using cross-entropy loss to supervise the predicted objectness probability. For mask generation, we employ both dice loss $\mathcal{L}_{dice}$ and pixel-wise cross-entropy loss $\mathcal{L}_{pix}$. The former improves shape consistency by maximizing the intersection-over-union (IoU) between predicted and ground-truth masks, while the latter refines pixel-level segmentation details. All loss components operate on sigmoid-activated outputs. Focal loss and dice loss use summation reduction, while cross-entropy losses use mean reduction.

\section{Experimental Results}
\label{sec:4}

\subsection{Study Areas}

Our PoplarDataset includes two poplar (Populus spp.) plots located in Huanghai Haibin National Forest Park, Dongtai, Jiangsu Province, China $(32^\circ 51^\prime N, 120^\circ 50^\prime E)$, as shown in Fig. \ref{fig:7}(a). Plot 1 spans approximately 10,000 $m^2$, featuring 12-year-old mature poplar trees that are densely distributed, with complex branching structures and lush canopies. Plot 2 covers about 3,600 $m^2$ and consists of 3-year-old juvenile poplars, with a diameter at breast height (DBH) ranging from 3 to 8cm, characterized by sparser branches. These two plots differ significantly in terms of tree density, DBH distribution, and trunk morphology, providing an ideal experimental platform and valuable data resources for analyzing poplar growth dynamics and phenotypic traits across developmental stages.

Data collection was conducted using a DJI M350RTK drone equipped with a P1 camera (Fig. \ref{fig:7}(b)), flying at an altitude of 5m along a predefined route to capture high-resolution aerial imagery. For data annotation, we used LabelMe to perform fine-grained labeling of each tree, including trunk and branch contours, and to generate instance-level masks and bounding box information.

Unlike traditional forestry datasets that primarily focus on mature trees, our PoplarDataset breaks this limitation by placing greater emphasis on the growth of juvenile trees. Given their high plasticity in gene expression during early growth stages, juvenile trees play a crucial role in efficient breeding and early-stage phenotyping studies \cite{meyer2018inter, lloyd2022epigenome}.

\subsection{Datasets}

\begin{figure}[h]
	\centering
	\includegraphics[width=\linewidth]{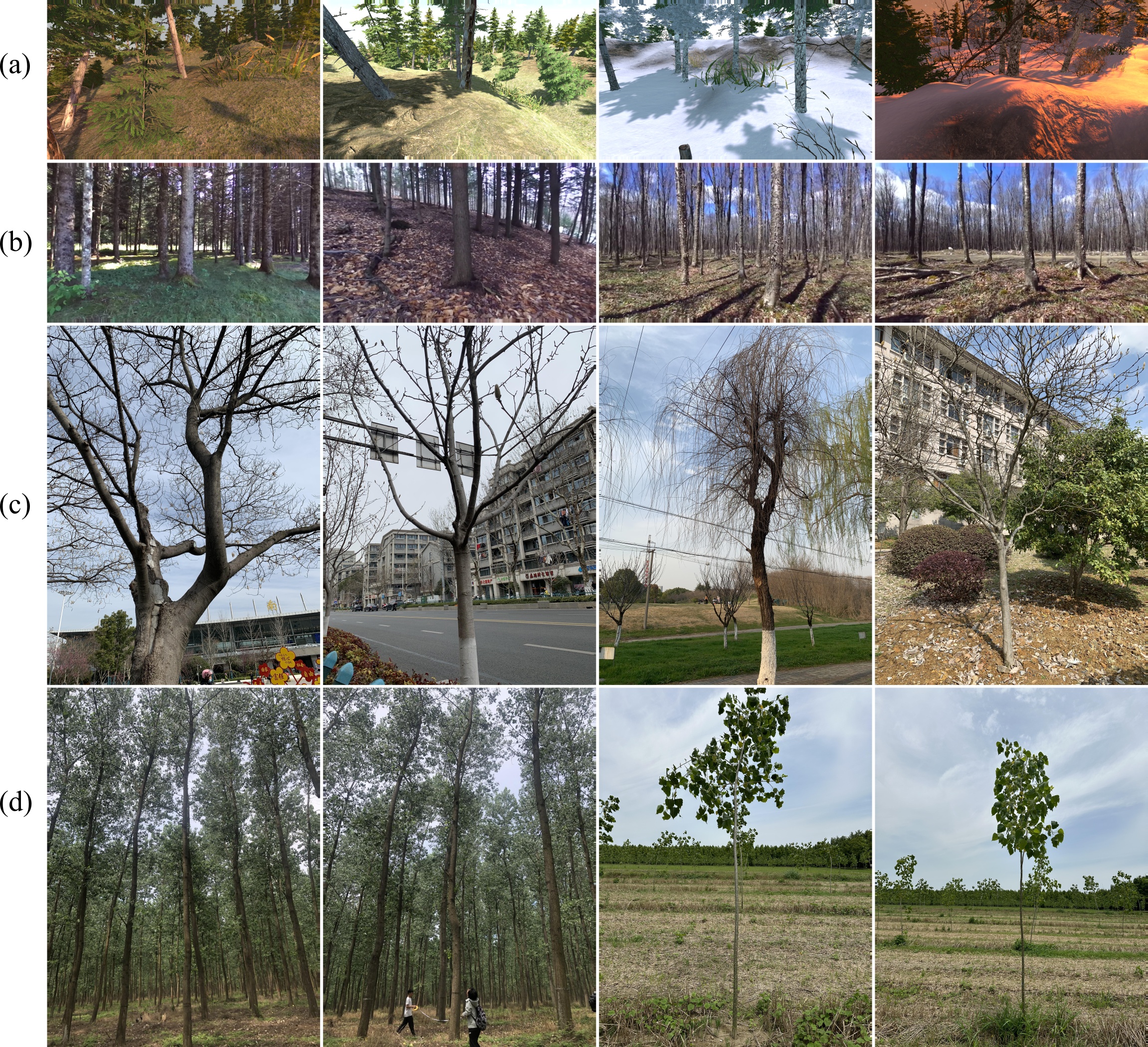}
	\caption{Samples from different datasets. (a) SynthTree43k; (b) CaneTree100; (c) Urban Street; (d) PoplarDataset.}
	\label{fig:8}
\end{figure}

This section introduces the four datasets used in experiments, aimed at further validating the generalization ability of our model in diverse environments. These datasets include synthetic and real images, covering multiple complex scenarios such as street trees, and natural forests, with sample illustrations shown in Fig. \ref{fig:8}.

\subsubsection{PoplarDataset}

is a field-collected dataset focusing on poplar (Populus spp.) trees at different growth stages, sourced from Huanghai Haibin National Forest Park in Jiangsu, China. The dataset contains 99 high-resolution images captured by a DJI M350RTK drone equipped with a P1 camera at a height of 5 meters. It covers two plots with distinct age structures—12-year-old mature trees and 3-year-old juvenile trees. Fine-grained instance annotations for 317 mature trees and 50 juvenile trees, including trunks and branches, were created using LabelMe. PoplarDataset is unique in its systematic documentation of juvenile tree morphology, addressing the gap in early-stage growth datasets. By capturing the phenotypic variability and genetic plasticity of juvenile trees, it provides essential support for research in precision breeding and forest growth modeling.

\subsubsection{SynthTree43k}

is proposed to address the gap in large-scale annotated datasets for forest environments, aiming to include as many realistic conditions as possible while avoiding manual annotation \cite{grondin2022tree}. The dataset consists of over 43k synthetic images—40k for training, 1k for validation, and 2k for testing—along with more than 190k annotated trees, generated through the Unity engine to simulate diverse tree models, environmental conditions, and lighting variations. Each image is annotated with bounding boxes, segmentation masks, and keypoints, providing comprehensive data for forest-related vision tasks.

\subsubsection{CaneTree100}

is collected from public, private, and commercial forests in Quebec, Canada. It consists of 100 RGB images and 100 depth images, with annotations for over 920 trees \cite{grondin2022tree}. The images were captured from 33 videos filmed with a handheld ZED stereo camera, with HD resolution of 1280 $\times$ 720 @ 60fps. The data was collected between June 2020 and April 2021, during the hours of 9 AM to 6 PM, ensuring a variety of lighting conditions. The dataset includes typical tree species such as fir, spruce, pine, birch, and maple, representing the ecological diversity of Canadian forests.

\subsubsection{Urban Street}

is a large-scale dataset for urban street tree recognition, collected from 10 cities across subtropical and temperate monsoon climate zones in China, covering 50 common tree species \cite{yang2023urban}. It contains 41,467 high-resolution images, including 22,872 with pixel-level polygon annotations of organs such as leaves, trunks, full trees, branches, flowers, and fruits, under diverse seasonal, lighting, and street scene conditions. This study uses the segmentation-branch subset focused on winter defoliation, which includes 1,485 branch images from 13 species with instance-level annotations of trunks and branch contours.

\subsection{Evaluation Metrics}

In this study, we adopt the standard MS COCO evaluation metrics to assess instance segmentation performance. Specifically, we utilize mask-based metrics to evaluate the quality of segmentation predictions, including Average Precision at various thresholds—namely, $mAP$, $AP50$, and $AP75$—as well as Average Recall ($AR$). These metrics comprehensively reflect the model’s accuracy and robustness across different Intersection over Union ($IoU$) thresholds.

\textbf{Intersection over Union} $IoU$ is a key metric for measuring the overlap between predicted segmentation results and ground truth annotations. It is defined as the ratio of the intersection area between the predicted mask $Pred$ and the ground truth mask $GT$ to their union area. The calculation formula is as follows:
\begin{align}
	IoU = \frac{Pred\cap GT}{Pred\cup GT}
\end{align}
The closer the $IoU$ value is to 1, the higher the overlap between the prediction and the ground truth, indicating better segmentation quality.

\textbf{Average Precision} $AP$ is used to evaluate a model's detection and segmentation performance at a specific $IoU$ threshold, based on the Precision-Recall (PR) curve. Precision and recall are defined as:
\begin{align}
	Precision=\frac{TP}{TP+FP}
\end{align}
\begin{align}
	Recall=\frac{TP}{TP+FN}
\end{align}
where $TP$, $FP$, and $FN$ represent True Positives, False Positives, and False Negatives, respectively.

In the COCO evaluation standard, $mAP$ is calculated by averaging $AP$ values across multiple $IoU$ thresholds (0.50:0.05:0.95). $AP50$ and $AP75$ represent $AP$ at fixed $IoU$ thresholds of 0.50 and 0.75, respectively.

\begin{table*}[h]
	\centering
	\caption{Comparison with previous methods on SynthTree43k-test and CaneTree100-test datasets.}
	\label{tab:quantitative1}
	\renewcommand{\arraystretch}{1.2}
	\rmfamily
	
	\begin{tabular}{cccccccccc}
		\hline
		\multirow{2}{*}{Method} & \multicolumn{4}{c}{SynthTree43k} & & \multicolumn{4}{c}{CaneTree100} \\ \cline{2-5}\cline{7-10}
		& $mAP$ & $AP50$ & $AP75$ & $AR$ & & $mAP$ & $AP50$ & $AP75$ & $AR$ \\ \hline
		Mask R-CNN & 26.1 & 65.2 & 14.0 & 38.0 & & 49.5 & 79.5 & 54.9 & 55.1 \\
		YOLACT & 28.9 & 58.8 & 24.9 & 41.1 & & 22.4 & 46.7 & 19.1 & 40.2 \\
		YOLO11 l-seg & 45.3 & 87.0 & 42.3 & 54.3 & & 36.0 & 65.5 & 37.2 & 50.1 \\
		QueryInst & 33.7 & 81.5 & 17.8 & 44.0 & & 38.7 & 61.7 & 45.3 & 44.9 \\
		SparseInst & 48.1 & 86.4 & 47.8 & 51.2 & & 42.2 & 74.8 & 46.3 & 49.6 \\
		FastInst & 51.1 & 89.7 & 51.2 & 55.1 & & 36.7 & 66.0 & 35.6 & 46.6 \\
		MambaVision & 31.9 & 73.1 & 18.8 & 41.6 & & 51.7 & 80.9 & 58.6 & 58.6 \\
		RF-DETR-Seg & 49.6 & 91.9 & 50.0 & 57.9 & & 46.3 & 81.7 & 46.7 & 56.8 \\
		\textbf{WaveInst(ours)} & \textbf{50.5} & \textbf{87.7} & \textbf{51.1} & \textbf{53.5} & & \textbf{47.0} & \textbf{78.0} & \textbf{53.2} & \textbf{53.8} \\ \hline
	\end{tabular}
\end{table*}

\begin{table*}[h]
	\caption{Comparison with previous methods on PoplarDataset-val.}
	\label{tab:quantitative2}
	\centering
	\renewcommand{\arraystretch}{1.2}
	\rmfamily
	
	\begin{tabular}{ccccccccccccc}
		\hline
		\multirow{2}{*}{Method} & \multicolumn{4}{c}{All} & & \multicolumn{3}{c}{Mature} & & \multicolumn{3}{c}{Juvenile} \\ \cline{2-5}\cline{7-9}\cline{11-13}
		& $mAP$ & $AP50$ & $AP75$ & $AR$ & & $mAP$ & $AP50$ & $AP75$ & & $mAP$ & $AP50$ & $AP75$ \\ \hline
		Mask R-CNN & 24.5 & 52.0 & 22.4 & 32.3 & & 44.4 & 73.3 & 44.9 & & 4.5 & 30.7 & 0.0 \\
		YOLACT & 13.6 & 43.8 & 10.9 & 25.1 & & 18.1 & 43.8 & 10.9 & & 9.2 & 66.3 & 0.0 \\
		YOLO11 l-seg & 16.1 & 78.0 & 2.7 & 30.6 & & 19.8 & 56.0 & 5.5 & & 12.4 & 100.0 & 0.0 \\
		QueryInst & 21.8 & 55.0 & 19.6 & 29.3 & & 39.2 & 72.5 & 39.1 & & 4.3 & 37.6 & 0.0 \\
		SparseInst & 32.5 & 85.4 & 27.2 & 37.5 & & 50.6 & 80.7 & 54.3 & & 14.4 & 90.1 & 0.0 \\
		FastInst & 30.6 & 72.6 & 26.7 & 36.7 & & 51.7 & 84.9 & 53.5 & & 9.5 & 60.4 & 0.0 \\
		MambaVision & 21.2 & 44.4 & 18.4 & 31.1 & & 40.9 & 72.6 & 36.8 & & 1.6 & 16.2 & 0.0 \\
		RF-DETR-Seg & 38.1 & 92.9 & 31.7 & 47.1 & & 53.7 & 85.9 & 63.5 & & 22.5 & 100.0 & 0.0 \\
		\textbf{WaveInst(ours)} & \textbf{41.1} & \textbf{89.5} & \textbf{31.1} & \textbf{46.4} & & \textbf{53.1} & \textbf{79.0} & \textbf{62.3} & & \textbf{29.1} & \textbf{100.0} & \textbf{0.0} \\ \hline
	\end{tabular}
\end{table*} 

\textbf{Average Recall} $AR$ is closely related to Average Precision $AP$ and aims to measure the model's detection ability across multiple recall thresholds. A higher recall indicates that the model is able to detect more actual targets. The calculation of $AR$ typically considers several different recall thresholds, and can be expressed by the following formula:
\begin{align}
	AR=\frac{1}{N}\sum_{i=1}^{N}{Recall_i}
\end{align}
where $Recall_i$ denotes the recall rate at the $i$-th $IoU$ threshold, and $N$ denotes the number of different recall thresholds considered.

\subsection{Implementation Details}

All experiments were conducted on a high-performance platform with an AMD Ryzen 9 7950X CPU and an NVIDIA RTX 3090 GPU (24 GB VRAM) running Ubuntu 22.04. The software environment included CUDA 11.8 and PyTorch 2.0.0. Training and inference were implemented using MMDetection, Detectron2, and Ultralytics frameworks.

For the SynthTree43k dataset, we initialize the model with pre-trained weights from ImageNet. For networks using ResNet50 as the backbone, we freeze the shallow layers (stem and stage 1) to preserve the ability to extract low-level features. To achieve better generalization on this synthetic dataset, we apply data augmentation techniques, including random adjustments to brightness, contrast, hue, and saturation, as well as the addition of Gaussian noise, ISO noise, and glass motion blur. The models are trained with the SGD optimizer (momentum 0.9, weight decay 1e-4, initial learning rate 0.005, batch size 4) for 36 epochs, with learning rate decaying by 0.1 at epochs 24 and 33.

We then perform transfer learning on three representative real-world datasets—CaneTree100, Urban Street, and PoplarDataset—which correspond to natural forest, urban forest, and artificial forest scenarios, respectively. Because the data scale in these datasets is relatively small, we switch to the AdamW optimizer, reduce the batch size to 2, and adopt a cosine annealing learning-rate schedule. The initial learning rate for all three datasets is set to 1e-4.

\subsection{Comparisons with Previous Methods}

We compare WaveInst with several state-of-the-art (SOTA) methods, including one-stage detectors, two-stage detectors, and query-based approaches. Through both quantitative and qualitative analysis, we demonstrate the advantages of our method over Mask R-CNN \cite{he2017mask}, YOLACT \cite{bolya2019yolact}, YOLO11 \cite{khanam2024yolov11}, QueryInst \cite{fang2021instances}, SparseInst \cite{cheng2022sparse}, FastInst \cite{he2023fastinst}, MambaVision \cite{hatamizadeh2025mambavision}, and RF-DETR \cite{robinson2025rf}.

\subsubsection{Quantitative Comparisons}

By introducing a high-frequency compensation branch based on discrete wavelet transform and an adaptive gated fusion module, the model effectively embeds the local high-frequency details lost in the backbone network into the feature maps output by the encoder, providing richer feature representations for the decoder. Meanwhile, the DRMask branch in the decoder utilizes the DySample module to efficiently reconstruct high-resolution feature maps, thereby significantly improving the segmentation accuracy of the model.

Table~\ref{tab:quantitative1} presents the performance with SOTA methods on the SynthTree43k-test and CaneTree100-test datasets. On the SynthTree43k-test dataset, our model achieves an $mAP$ of 50.5, closely trailing the top-performing FastInst at 51.1, while outperforming RF-DETR-Seg (49.6) and SparseInst (48.1), showcasing strong competitiveness. On the real-world forest scene dataset, CaneTree100-test, MambaVision ranks first with an $mAP$ of 51.7, ahead of Mask R-CNN’s 49.5. However, their performance on SynthTree43k-test is terrible, indicating instability. In contrast, our method and RF-DETR-Seg achieve more stable results on CaneTree100 with $mAP$ of 47.0 and 46.3, respectively.

\begin{figure*}[p]
	\centering
	\includegraphics[width=0.8\textwidth]{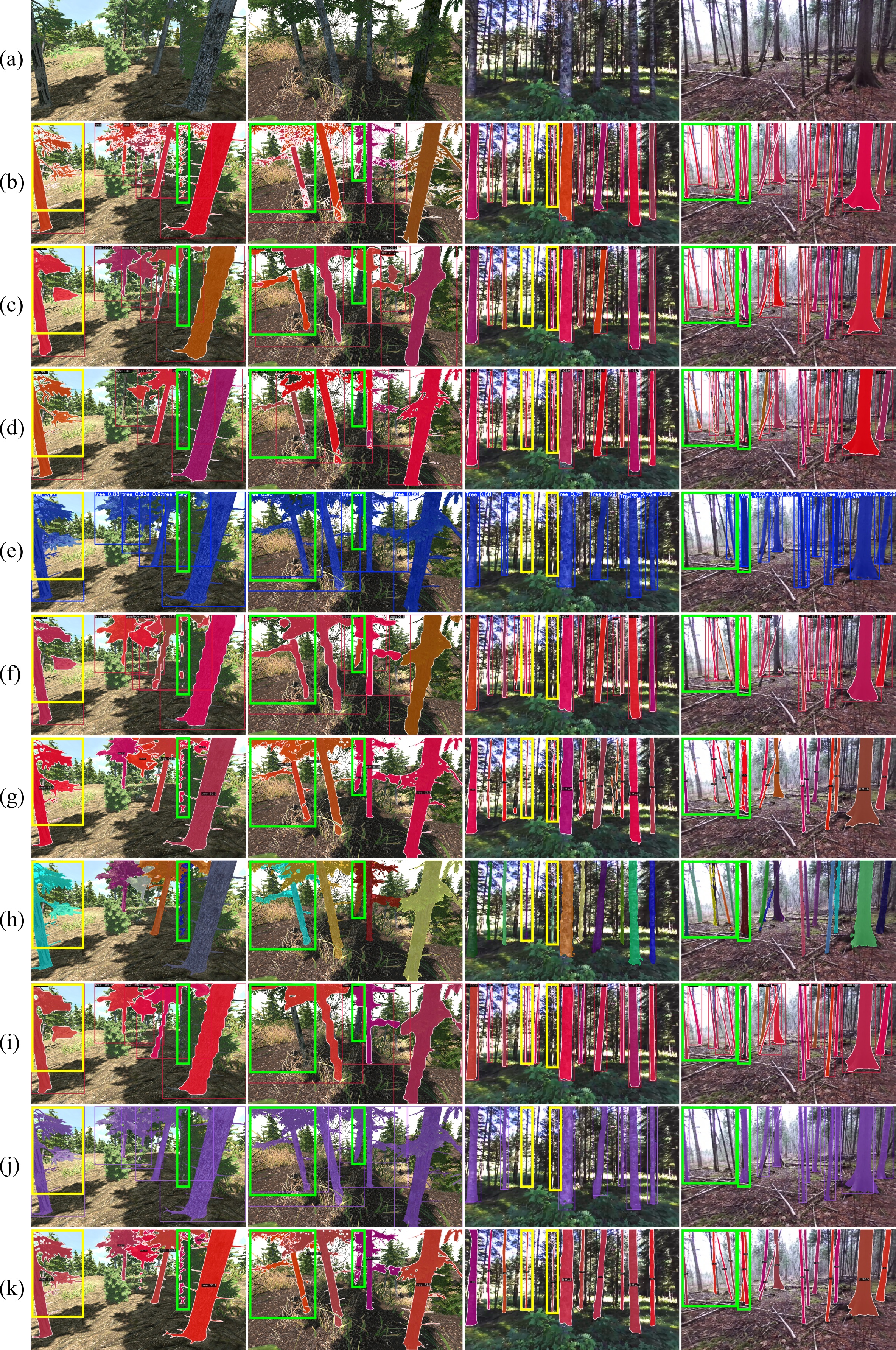}
	\caption{Visualization results on SynthTree43k and CaneTree100. (a) input image; (b) ground truth; (c) Mask R-CNN; (d) YOLACT; (e) YOLO11 l-seg; (f) QueryInst; (g) SparseInst; (h) FastInst; (i) MambaVision; (j) RF-DETR-Seg; (k) WaveInst (ours).}
	\label{fig:9}
\end{figure*}

\begin{figure*}[h]
	\centering
	\includegraphics[width=0.8\textwidth]{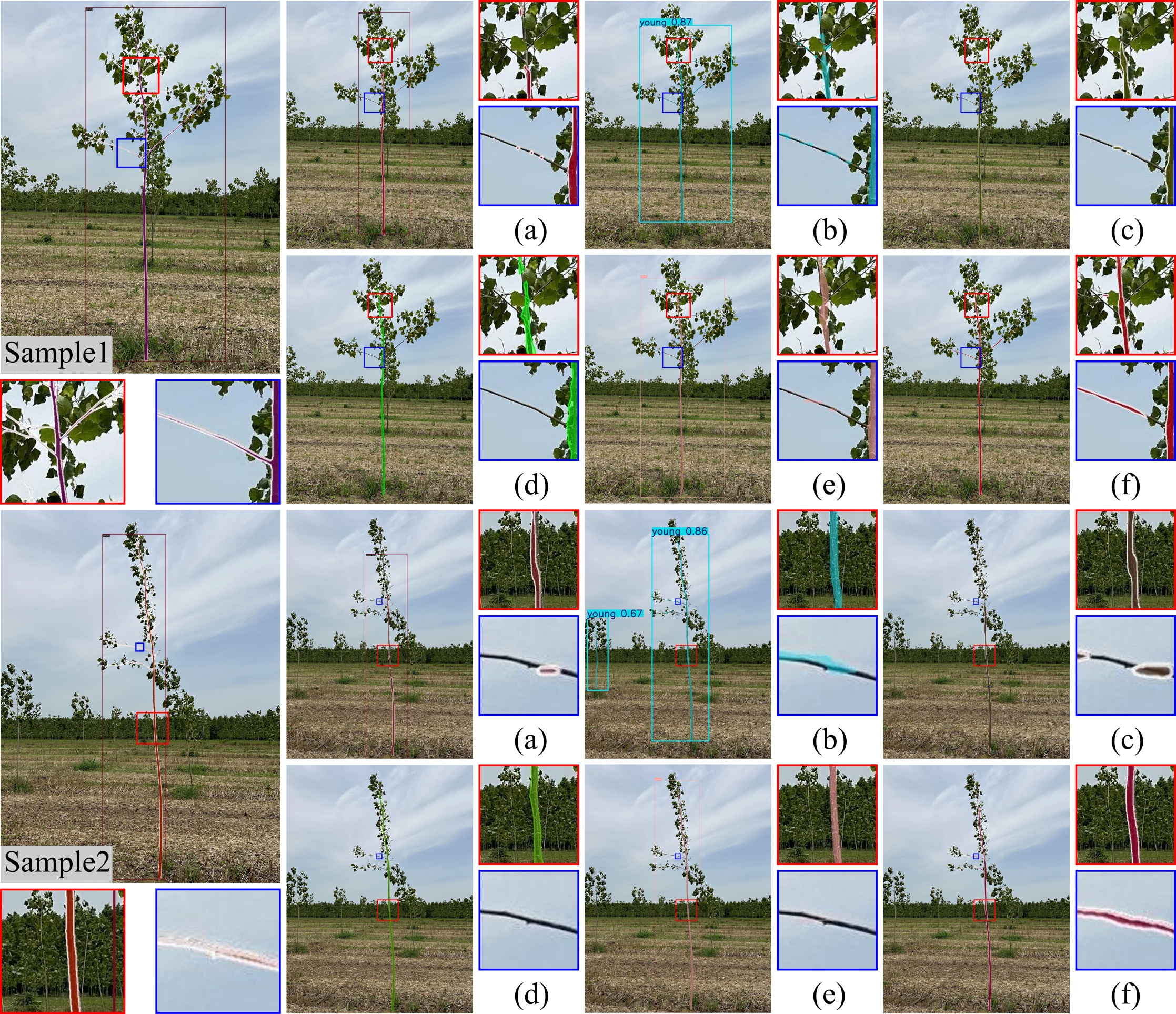}
	\caption{Visualization results on PoplarDataset(Juvenile). (a) YOLACT; (b) YOLO11 l-seg; (c)   SparseInst; (d) FastInst; (e) RF-DETR-Seg;  (f) WaveInst (ours).}
	\label{fig:10}
\end{figure*}

Table~\ref{tab:quantitative2} presents the quantitative results of the models on the PoplarDataset-val, which contains both mature and juvenile trees. Our proposed WaveInst achieves an overall $mAP$ of 41.1, surpassing the best-preforming RF-DETR-Seg (38.1) by 3.0 points and our baseline model SparseInst (32.5) by 8.6 points, validating the effectiveness of the frequency-domain enhancement in boosting segmentation accuracy. In the mature category, RF-DETR-Seg achieves a slightly higher $mAP$ of 53.7, with WaveInst following closely at 53.1. Except for YOLACT and YOLO11 l-seg, the other performances are acceptable. However, in the more challenging juvenile category, characterized by thin and structurally complex trunks, the performance of most methods drops significantly; for instance, MambaVision achieves only 1.6, and FastInst reaches 9.5. In contrast, WaveInst attains an $mAP$ of 29.1, improving upon RF-DETR-Seg (22.5) by 6.6 points. Although three methods achieve a perfect $AP50$ of 100.0 in this category, none of the methods score on $AP75$.

For the juvenile category annotations, we not only labeled the main trunk but also included the small branches. A rough estimation shows that each juvenile tree has around 10 branches, with pixel sizes often only in the single digits. Since the mask area of these fine branches accounts for about $28\%$ of the total annotated area, their fine details pose a significant challenge for the segmentation algorithm. As a result, despite several models achieving perfect scores in $AP50$, all models scored 0.0 in $AP75$, indicating that current segmentation algorithms still face considerable difficulty when handling these small, low-resolution branches.

The performance difference mainly arises from how model architectures adapt to data scale and distribution. FastInst uses a hybrid design with a CNN-based backbone and pixel decoder for multi-scale feature extraction, while its core instance decoder employs a dual-path transformer with instance activation-guided queries for global modeling. This enables FastInst to leverage the Transformer’s long-range dependency capture on large datasets like SynthTree43k, but struggles with smaller, complex datasets like CaneTree100 \cite{akkaya2024enhancing}. Similarly, MambaVision combines residual convolution blocks for local features with an state space model-based Mamba Mixer for global modeling and Transformer blocks for context restoration. However, its complexity makes it sensitive to data distribution, excelling only on CaneTree100 and underperforming on SynthTree43k and PoplarDataset. RF-DETR-Seg, using a unified transformer architecture, offers stable, strong performance across all datasets due to its global modeling ability. In contrast, our WaveInst relies solely on a CNN-based architecture, avoiding the dependency on large-scale pretraining. CNN’s local inductive bias allows stable and efficient learning on smaller datasets like CaneTree100, while the frequency-domain branch improves instance boundary perception, leading to exceptional performance on the juvenile class in PoplarDataset.
 
\begin{figure*}[h]
	\centering
	\includegraphics[width=0.8\textwidth]{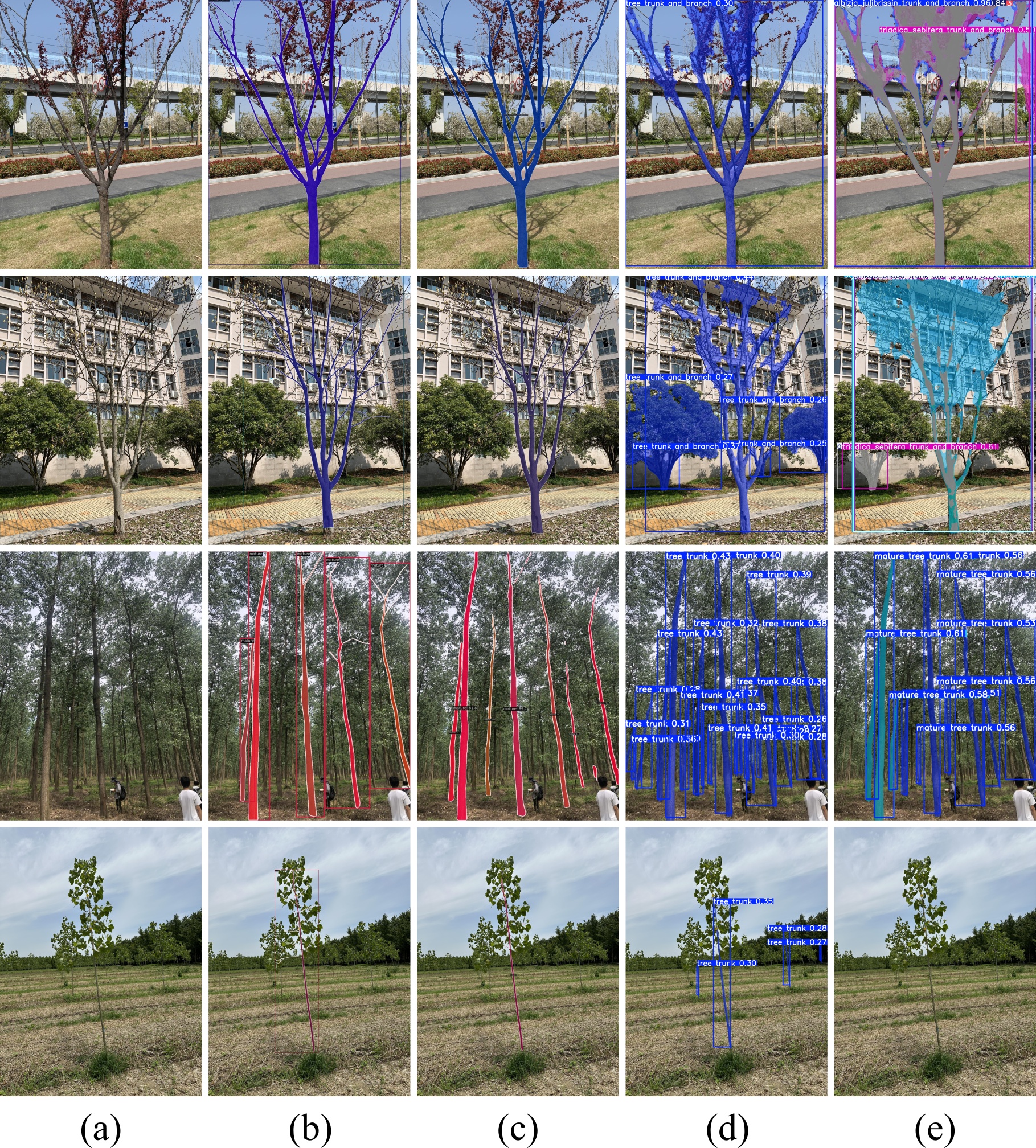}
	\caption{Qualitative comparison with SAM3. (a) input image; (b) ground truth; (c) WaveInst(outs); (d) SAM3+single prompt; (e) SAM3+multiple prompts.}
	\label{fig:11}
\end{figure*}
 
\subsubsection{Qualitative Comparisons}

Fig. \ref{fig:9} presents a comparative visualization of segmentation results from different models on the SynthTree43k and CaneTree100 datasets. The green boxes highlight segmentation performance under occlusions that cause incomplete and discontinuous structures, while the yellow boxes focus on segmentation accuracy in finer details. In column 1, for severely occluded regions, only QueryInst, SparseInst, FastInst (f-h), and WaveInst (k) successfully segmented the targets, though the first two methods produced fragmented results. In column 2, under slight occlusion, Mask R-CNN, YOLACT, and YOLO11 l-seg (c-e) missed small targets, and MambaVision missed both small and some larger targets. In real-world scenarios, for distant and thin trunks shown in columns 3 and 4, other methods exhibited varying levels of missed detections or discontinuous segmentation, with RF-DETR-Seg (j)—despite having the highest $mAP$—suffering from more significant missed detections. In contrast, our method showed better stability and finer detail handling. In column 1, for handling leaf edges and gaps on the left, YOLACT (d), SparseInst (g), RF-DETR-Seg (j), and WaveInst (k) outperformed other methods in terms of detail preservation. In column 4, under the interlacing condition, other models either failed to distinguish the overlapping trunks or merged them into a single object, while our model handled this scenario more effectively.

In Fig. \ref{fig:10}, we compare the performance of multiple methods on the segmentation of juvenile trees. In the trunk area, sample 1 shows the finer regions near the top. YOLACT, YOLO11 l-seg, SparseInst, and FastInst (a-d) fail to generate complete masks. Although RF-DETR-Seg (e) produces a more complete mask, its boundary segmentation accuracy is slightly inferior compared to our WaveInst (f). Sample 2 demonstrates the segmentation results in the middle of the trunk, where all methods show significant improvement in completeness, but YOLACT (a) and SparseInst(c) still exhibit substantial segmentation errors. For the fine branches, only WaveInst successfully performs the segmentation in sample 1. Although it does not form a fully continuous connection with the main trunk, WaveInst's performance stands out compared to the other methods, which fail to segment the fine branches at all. In sample 2, although YOLO11 l-seg generates more masks, there is a significant gap between its results and the ground truth, whereas WaveInst performs relatively better. Overall, despite RF-DETR-Seg achieving an $mAP$ of 22.5 in the juvenile category, its ability to segment fine branches is inferior to WaveInst, which performs better in fine branch segmentation. While WaveInst still has room for improvement in connecting the main trunk with the fine branches, its overall performance surpasses that of RF-DETR-Seg.

\begin{figure*}[h]
	\centering
	\includegraphics[width=0.8\textwidth]{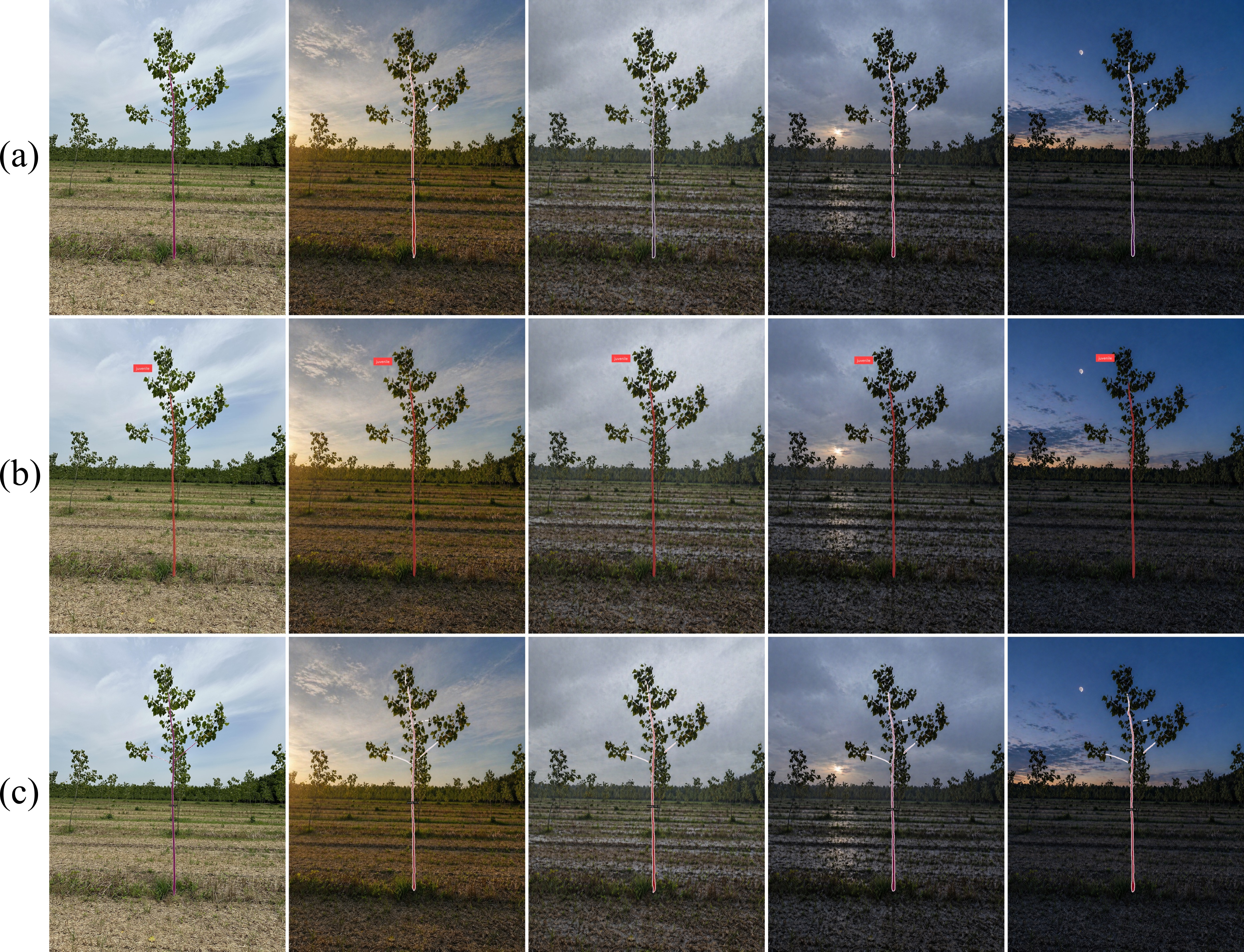}
	\caption{Robustness under varying lighting conditions. (a) SparseInst; (b) RF-DETR-Seg; (c) WaveInst (ours).}
	\label{fig:16}
\end{figure*}

In addition, recent research has shifted towards general-purpose vision foundation models, such as the SAM series. Specifically, the latest version, SAM3, introduces a new feature that allows visual concepts to be specified via text prompts \cite{carion2025sam}. In Fig. \ref{fig:16}, we present a qualitative comparison between the proposed WaveInst and SAM3 for tree trunk segmentation to further demonstrate the advantages of our method. For SAM3, we tested both single and multiple prompt approaches, where single prompt does not distinguish between tree species or growth stages, and multiple prompts do. In the two sample images from the Urban Street dataset, SAM3's segmentation accuracy is significantly lower than that of our method, and in the case of using multiple prompts, mask overlap occurs. Although we raised the label confidence threshold from 0.25 in single prompt to 0.5, the mask overlap issue still persists, indicating that SAM3 has insufficient capability to distinguish tree species information. Similarly, in the PoplarDataset, SAM3 with multiple prompts generates overlapping masks for the same tree, one labeled as mature and the other as juvenile. However, despite this, SAM3 does manage to segment the trees in the foreground while also incorrectly segmenting distant, unannotated trees, which is something our model does not achieve. Overall, while SAM3 has certain strengths, our model performs better in terms of precision and fine detail handling.

In real forestry scenarios, besides the challenges of overlapping trees in SynthTree43k and CaneTree100, models must be robust to varying lighting conditions. We simulate different illumination scenarios on the PoplarDataset and evaluate the top three models. As shown in Fig. \ref{fig:16}, SparseInst (a) produces unsmooth contours and degrades under lighting changes; RF-DETR-Seg (b) generates smooth masks but shows poor alignment with branches and limited fine-branch capture; in contrast, our method maintains stable segmentation across lighting conditions, with moderate smoothness, highest structural alignment, and strongest branch capture, demonstrating superior overall robustness.

\subsection{Ablation Study}
\label{sec:ablation}
\subsubsection{DWT Block}

In this section, we evaluate the effectiveness of the proposed DWT Block, which utilizes the discrete wavelet transform to decompose feature maps into frequency domains. The configuration of different wavelet bases has a notable impact on the model's performance \cite{hajiabadi2021comparison}. To assess the effectiveness of various configurations, we conduct a series of experiments comparing both the order of wavelet (the number of vanishing moments) and the type of wavelet basis used in the decomposition.

\subsubsection*{a. Order of wavelet basis}

\begin{figure}[h]
	\centering
	\includegraphics[width=\linewidth]{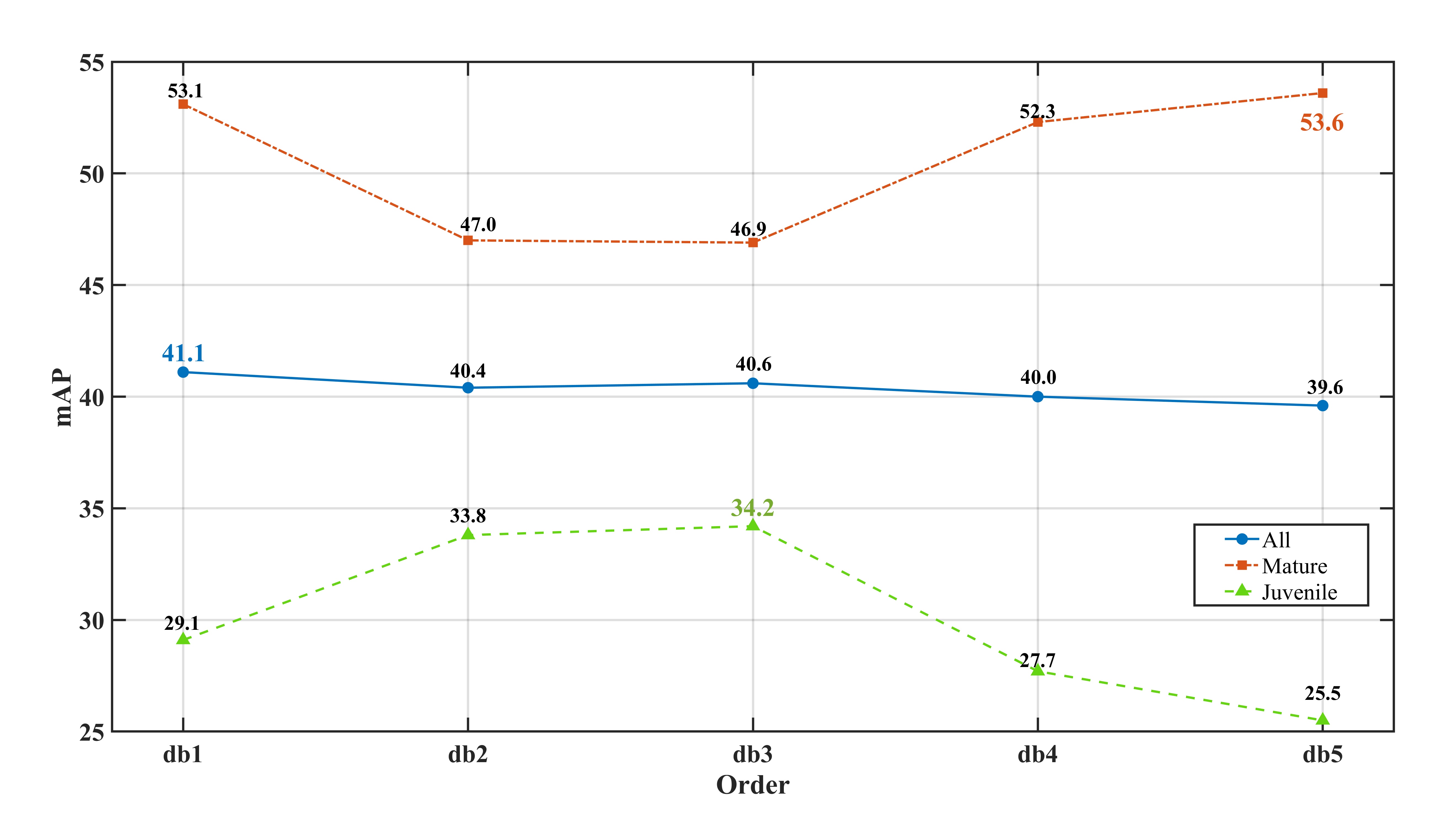}
	\caption{Impact of wavelet order on PoplarDataset-val.}
	\label{fig:order}
\end{figure}

We evaluate the performance of the DWT Block using wavelets from the Daubechies family, specifically db1 through db5. As shown in Fig. \ref{fig:order}, on PoplarDataset, the segmentation performance for the Mature category initially decreases and then increases as the wavelet order progresses from db1 to db5, while the Juvenile category shows the opposite trend. We consider that this behavior arises from the significant differences in the truck scales between the two growth stage of the trees, which cause competition in the network's ability to learn high-frequency features extraction from different scales. Overall, db1 shows the best segmentation performances, suggesting that lower-order wavelets are sufficiently effective in capturing the frequency information.

\subsubsection*{b. Type of wavelet basis}

\begin{figure}[h]
	\centering
	\includegraphics[width=\linewidth]{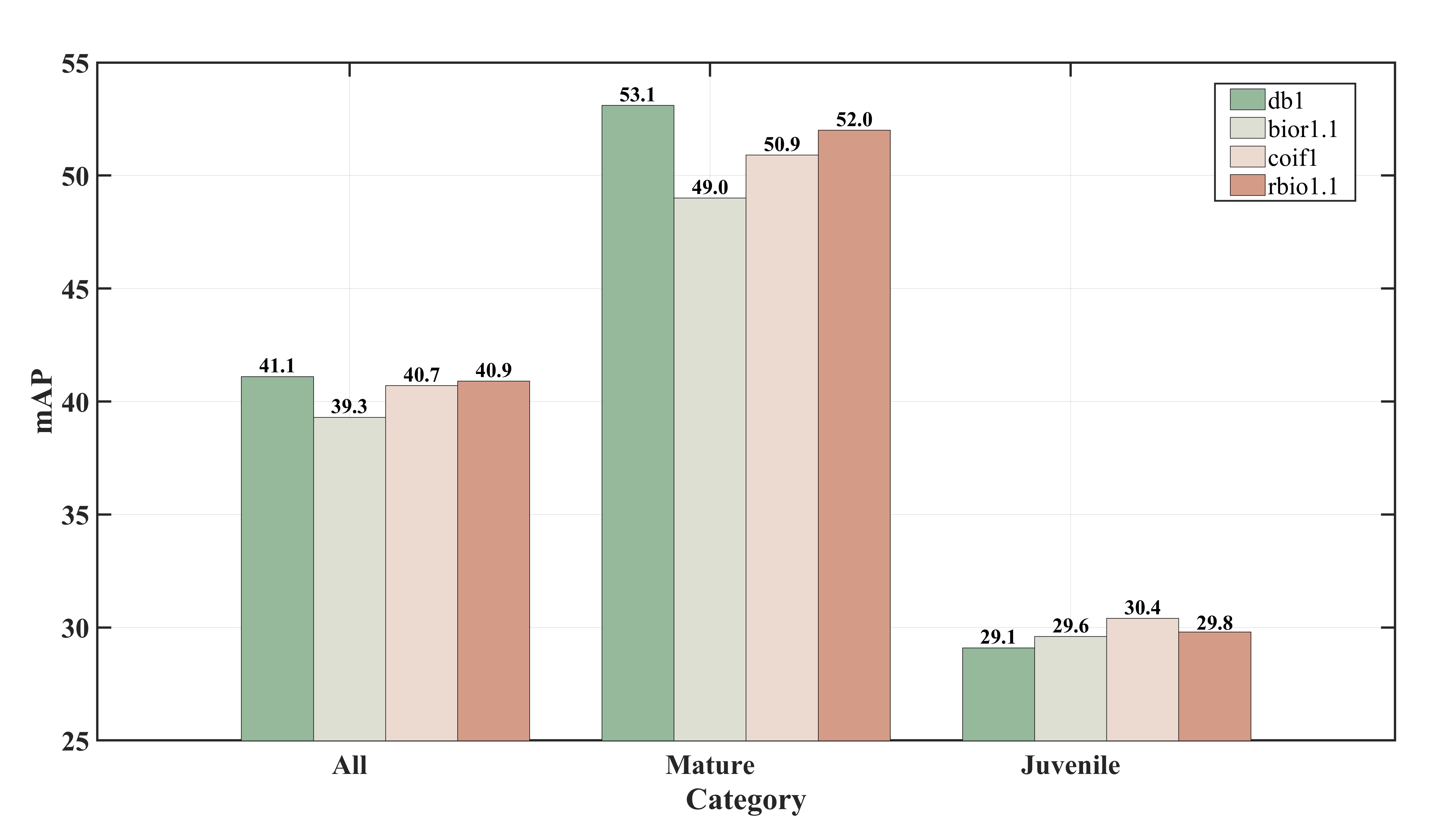}
	\caption{Impact of wavelet type on PoplarDataset-val.}
	\label{fig:type}
\end{figure} 

Furthermore, we compare db1 with lower-order wavelet functions from the Biorthogonal, Coiflet and Reverse Biorthogonal families to evaluate the impact of different wavelet bases on performance. As shown in Fig. \ref{fig:type}, the segmentation performance across these wavelet bases is similar. Among them, db1 and rbio1.1 perform better for the Mature category, while coif1 and rbio1.1 excel for the Juvenile category. Overall, db1 achieves the best performance across both categories.

\subsubsection{HFE Block}

\begin{table*}[h]
	\caption{Impact of HFE Block on PoplarDataset-val.}
	\label{tab:hfe}
	\centering
	\renewcommand{\arraystretch}{1.2}
	\rmfamily
	
	\begin{tabular}{cccccccccccc}
		\hline
		\multirow{2}{*}{Module} & \multicolumn{3}{c}{All} & & \multicolumn{3}{c}{Mature} & & \multicolumn{3}{c}{Juvenile} \\ \cline{2-4}\cline{6-8}\cline{10-12}
		& $mAP$ & $AP50$ & $AP75$ & & $mAP$ & $AP50$ & $AP75$ & & $mAP$ & $AP50$ & $AP75$ \\ \hline
		HFE & 41.1 & 89.5 & 31.1 & & 53.1 & 79.0 & 62.3 & & 29.1 & 100.0 & 0.0 \\
		SE & 39.5 & 89.1 & 27.2 & & 48.9 & 78.2 & 54.3 & & 30.0 & 100.0 & 0.0 \\
		CBAM & 37.7 & 85.2 & 25.9 & & 46.8 & 70.3 & 51.8 & & 28.5 & 100.0 & 0.0 \\
		ECA & 39.3 & 89.9 & 28.2 & & 50.2 & 79.8 & 56.4 & & 28.5 & 100.0 & 0.0 \\
		CA & 38.8 & 88.8 & 28.6 & & 49.6 & 77.6 & 57.2 & & 28.0 & 100.0 & 0.0 \\
		SimAM & 38.8 & 89.4 & 25.2 & & 47.9 & 78.9 & 50.4 & & 29.8 & 100.0 & 0.0 \\ \hline
	\end{tabular}
\end{table*}
 
\begin{table*}[!htbp]
	\caption{Ablation study on PoplarDatase-valt.}
	\label{tab:ablation}
	\centering
	\renewcommand{\arraystretch}{1.2}
	\rmfamily
	
	\begin{tabular}{ccccccccccccccc}
		\hline
		DWT & \multirow{2}{*}{AGFM} & \multirow{2}{*}{DySample} & \multicolumn{3}{c}{All} & & \multicolumn{3}{c}{Mature} & & \multicolumn{3}{c}{Juvenile} & Params. \\ \cline{4-6} \cline{8-10} \cline{12-14}
		Branch & & & $mAP$ & $AP50$ & $AP75$ & & $mAP$ & $AP50$ & $AP75$ & & $mAP$ & $AP50$ & $AP75$ & (M) \\
		\hline
		& & & 32.5 & 85.4 & 27.2 & & 50.6 & 80.7 & 54.3 & & 14.4 & 90.1 & 0.0 & 31.617 \\
		\checkmark & & & 36.8 & 89.8 & 26.5 & & 48.7 & 79.5 & 53.0 & & 25.0 & 100.0 & 0.0 & 32.036  \\
		\checkmark & \checkmark & & 37.5 & 89.9 & 28.3 & & 48.7 & 79.9 & 56.7 & & 26.3 & 100.0 & 0.0 & 32.053 \\
		& & \checkmark & 36.4 & 90.3 & 24.5 & & 48.2 & 80.5 & 48.9 & & 24.5 & 100.0 & 0.0 & 31.650  \\
		\checkmark & & \checkmark & 39.7 & 89.1 & 29.9 & & 51.2 & 78.3 & 59.7 & & 28.2 & 100.0 & 0.0 & 32.069 \\
		\checkmark & \checkmark & \checkmark & 41.1 & 89.5 & 31.1 & & 53.1 & 79.0 & 62.3 & & 29.1 & 100.0 & 0.0 & 32.086 \\
		\hline
	\end{tabular}
\end{table*}

In order to better integrate high-frequency prior information into the network, we propose the HFE Block. It is designed as a multi-path feature extraction and aggregation module, aimed at improving the representational power of neural networks by capturing essential high-frequency components. It bears similarities to several well-established attention mechanisms and feature aggregation modules, such as SE \cite{hu2018squeeze}, CBAM \cite{woo2018cbam}, ECA \cite{wang2020eca}, CA \cite{hou2021coordinate}, and SimAM \cite{yang2021simam}, all of which focus on refining feature selection by dynamically assigning different levels of importance to various parts of the input.

To validate the effectiveness of the proposed HFE Block, we conduct a series of experiments in which it is replaced by the aforementioned classic modules. The experimental results, as shown in Table~\ref{tab:hfe}, demonstrate that the HFE Block achieves a $mAP$ of 41.1 on PoplarDataset, outperforming several classic modules. While SE and SimAM perform better in the Juvenile category, their performance in the Mature category is relatively weaker. In contrast, the proposed HFE Block leverages multi-path feature extraction, consistently outperforming other methods across all metrics. It demonstrates a superior ability to capture high-frequency information and enhance feature aggregation more effectively.

\subsubsection{Overall}

\begin{figure}[h]
	\centering
	\includegraphics[width=\linewidth]{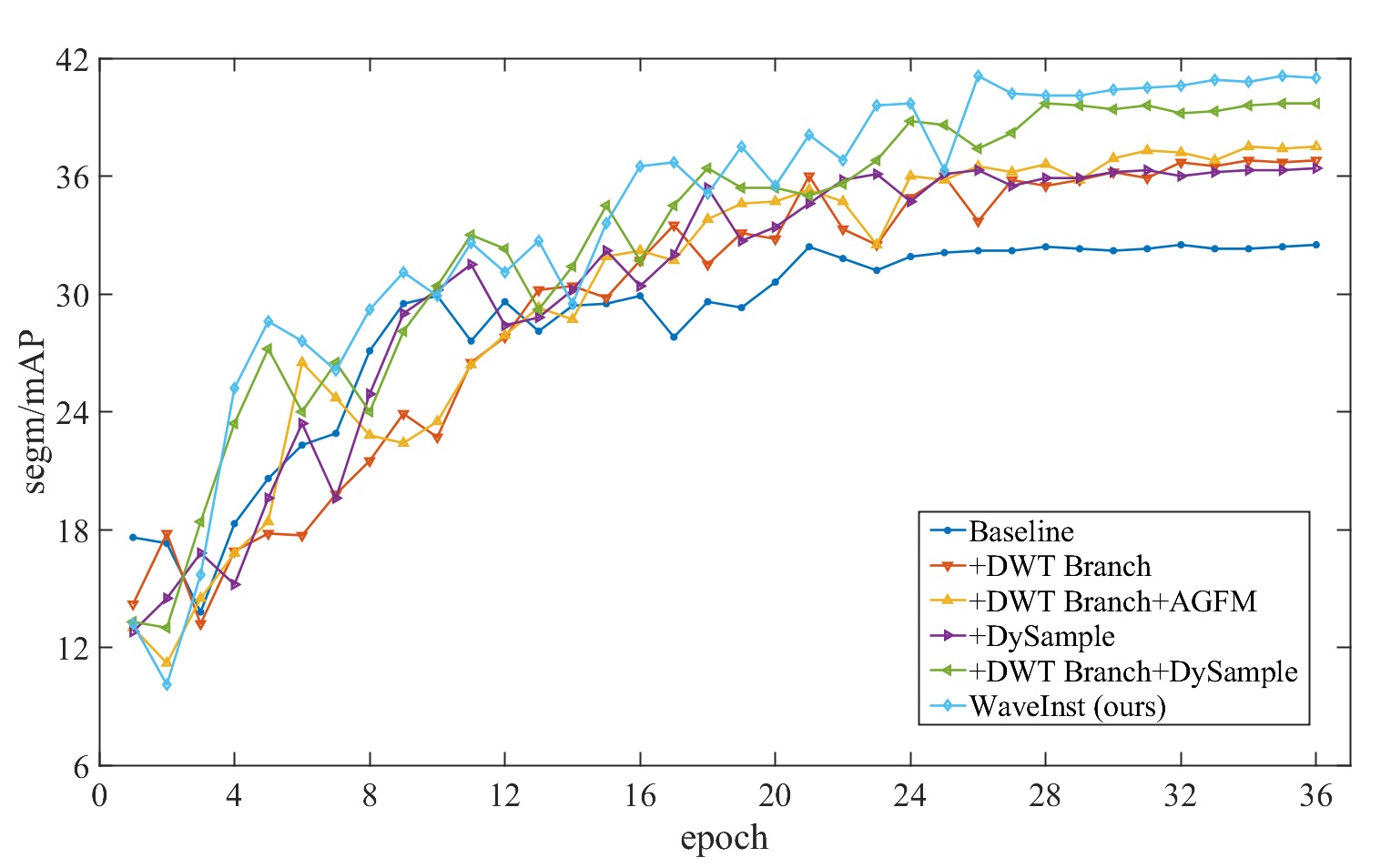}
	\caption{Training metric curves under different module combinations.}
	\label{fig:12}
\end{figure}

We conducted a series of ablation experiments to investigate WaveInst and verify the effectiveness of the proposed method. Table~\ref{tab:ablation} presents the segmentation performance under different module combinations. When AGFM is not used, the features from different branches are simply added together. The experimental results show that the high-frequency feature information embedded in the FFC branch effectively improves the segmentation accuracy for juvenile trees. Compared to simple feature summation, AGFM utilizes an adaptive gating mechanism to fully exploit the contribution of features from different branches, significantly enhancing feature fusion performance. DySample, through efficient feature reconstruction, enhances the network's ability to capture fine details. Additionally, we provide the training metric curves for the experiments in Fig. \ref{fig:12}, which visually demonstrate the training process.

\section{Discussion}
\label{sec:5}

\begin{figure*}[h]
	\centering
	\includegraphics[width=\textwidth]{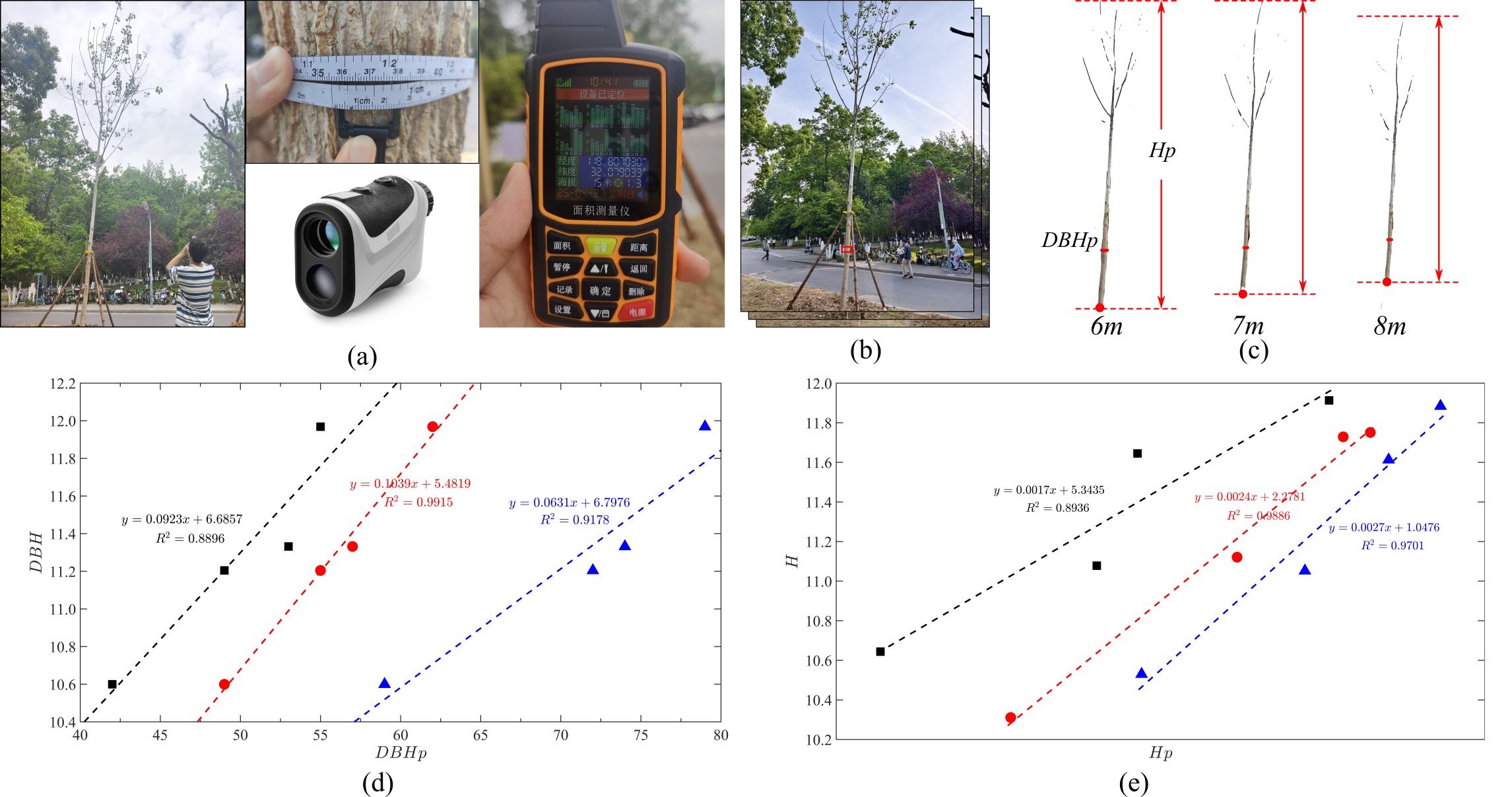}
	\caption{Estimation of tree growth parameters at different distance scales. (a) Measurement process and equipment used (diameter tape, rangefinder, GPS); (b) Collected images; (c) Segmentation results of our model; (d) DBH regression curves; (e) Tree height regression curves.}
	\label{fig:13}
\end{figure*}

This section discusses the performance and potential applications of the proposed WaveInst framework. By leveraging an instance segmentation strategy designed for thin tree trunks, WaveInst can effectively extract tree structural information from single RGB images. Compared with traditional measurement approaches, the proposed method simplifies the data acquisition process and reduces manual effort while maintaining reliable accuracy. The following subsections further analyze the efficiency and effectiveness of the proposed method compared with point cloud approaches, and discuss its applications in DBH and tree height estimation as well as tree species identification.

\subsection{Efficiency and Effectiveness}

We collect the inference runtimes of representative point cloud instance segmentation methods, including HAIS, SoftGroup, and ISBNet, from \cite{ngo2023isbnet}, and report the runtime of our model on the same Titan X GPU for comparison. These These point cloud methods are evaluated on the ScanNetV2 validation set, with each scene sampled to 200k points, while the proposed WaveInst is evaluated on the SynthTree43k-val set using RGB images of resolution 1280$\times$720.

\begin{table}[!htbp]
	\caption{Runtime comparison between point cloud and RGB instance segmentation methods.}
	\label{tab:runtime}
	\centering
	\renewcommand{\arraystretch}{1.2}
	\rmfamily
	
	\begin{tabular}{cccc}
		\hline
		Method & Modality & Scale & Runtime (ms)\\
		\hline
		HAIS & Point Cloud & 200k & 339 \\
		SoftGroup & Point Cloud & 200k & 345 \\
		ISBNet & Point Cloud & 200k & 237 \\
		WaveInst (ours) & RGB & 1280$\times$720 & 226 \\
		\hline
	\end{tabular}
\end{table}

As shown in Table~\ref{tab:runtime}, WaveInst achieves an inference time of 226ms, demonstrating a certain efficiency advantage over existing point cloud methods. This difference mainly arises because point cloud approaches typically require operations such as neighborhood search, sparse convolution, or clustering, which introduce considerable computational overhead when processing large-scale point sets. In contrast, image-based methods benefit from regular grid structures and highly optimized 2D convolution operations, resulting in lower computational cost. Moreover, ScanNetV2 mainly contains indoor scenes, while real-world forestry applications are typically outdoor environments with larger spatial scales and more complex structures, where point clouds may contain significantly more points, further increasing the computational burden of point cloud-based methods.

Point clouds have strong penetration ability and can capture three-dimensional information of the understory, providing a natural advantage in distinguishing overlapping trunks. However, point cloud acquisition and processing are costly and time-consuming, and for fine trunks, sparse sampling due to large inter-point spacing often results in missing data. In contrast, under-canopy orthophotos are inexpensive and easy to acquire, making them suitable for large-scale applications, but they lack depth information and cannot directly distinguish overlapping trunks, typically requiring multi-view observations to recover three-dimensional positions. To address these limitations, our method introduces a frequency-domain feature compensation branch that explicitly preserves and enhances high-frequency information, guiding the network to accurately delineate trunk boundaries, effectively mitigating segmentation confusion caused by overlapping trunks, and significantly improving the extraction of fine tree trunks.

\subsection{DBH and Height Estimation}

Diameter at Breast Height (DBH) and tree height are crucial indicators for assessing the growth of forest trees, reflecting the growth state and performance of different genotypes under specific environmental conditions. However, traditional DBH and tree height measurement methods primarily rely on manual operations, which are labor-intensive, inefficient, and prone to inconsistencies in data collection accuracy and standards \cite{laurin2019tree, wang2019field}. To overcome these challenges, we propose an automated measurement method combining instance segmentation and regression modeling, aiming to enhance the efficiency and accuracy of DBH and tree height measurements.

Specifically, we capture tree images using sensors and apply the proposed WaveInst for fine-grained segmentation. Subsequently, through regression modeling, we correlate the extracted trunk pixel width and pixel height with actual DBH and tree height data to establish models that automatically predict DBH and height from a single image. To verify the feasibility of this method, we conduct field experiments (as shown in Fig. \ref{fig:13}(a)) by capturing images of four poplar trees at different distance scales while simultaneously recording the corresponding DBH and tree height data. These data provide the foundation for establishing regression models. The experimental results show excellent fitting performance across different capture distances. As shown in Fig. \ref{fig:13}(d) and Fig. \ref{fig:13}(e), the $R^2$ values for DBH prediction are 0.8896, 0.9915, and 0.9178, while the $R^2$ values for tree height prediction are 0.8936, 0.9886, and 0.9701 at different capture distances.

Although jagged edges appear in Fig. \ref{fig:13}(c), they do not affect the parameter estimation. Taking the segmentation result at a distance of 6m as an example, the number of pixels at breast height $DBH_p$ is 82, corresponding to a predicted $DBH$ of 11.97cm. The ground truth pixel value is 74, corresponding to an actual $DBH$ of 11.33cm. The prediction error is approximately 1.06\% (less than 3\%), which is within an acceptable range. In future work, boundary smoothing modules (e.g., Sobel operators or Gaussian filtering) or optimized sampling strategies will be introduced to further alleviate such artifacts.

\subsection{Tree Species Recognition}

In addition, tree species information is equally valuable in tree structure analysis and forest conservation. The ability to identify tree species is crucial for accurately assessing the diversity and health of forest ecosystems \cite{yun2025lginet}. As an instance segmentation model focused on thin tree trunks, our method not only performs fine-grained tree segmentation but also possesses the ability to recognize tree species, which is demonstrated in Fig. \ref{fig:11}. The first row shows \textit{Prunus cerasifera f. atropurpurea}, while the second row shows \textit{Magnolia liliiflora Desr}. Despite the morphological similarities between these two species, our model effectively distinguishes them, showcasing its high precision and stability in tree species identification.

It should be noted that our test images do not include scenes captured under adverse weather or extreme lighting conditions, which would impose higher demands on target detection and tree species recognition. However, the SynthTree43k dataset provides synthetic images under various conditions, including nighttime, dusk, rainy, and snowy scenes for training, and data augmentation techniques such as color jittering are applied to enhance the model’s adaptability to different lighting and weather. In future work, we plan to further expand the test set with more real-world images captured under challenging conditions to comprehensively evaluate the robustness of the proposed method.

\section{Conclusion}
\label{sec:6}

This paper presents a novel instance segmentation network for individual tree structure extraction from different forests. The proposed method introduces a Frequency-domain Feature Compensation (FFC) Branch to enhance the perception of high-frequency texture information, and incorporates an Adaptive Gated Fusion Module (AGFM) to effectively fuse multi-scale semantic features with frequency-domain details. Experiments conducted on two representative forestry image datasets, SynthTree43k and CaneTree100, demonstrate that WaveInst outperforms existing  methods across multiple metrics, including $mAP$, $AP50$, and $AP75$. Furthermore, WaveInst exhibits superior fine-grained structural perception in challenging scenarios such as the collected Urban Street and PoplarDataset, addressing the current extraction limitations in understory detail recognition.

\appendices

\ifCLASSOPTIONcaptionsoff
  \newpage
\fi

\bibliographystyle{IEEEtran}
\bibliography{IEEEabrv,mylib}

\end{document}